\documentclass[10pt,twocolumn,letterpaper]{article}

\usepackage{cvpr}              % To produce the CAMERA-READY version
\usepackage[dvipsnames]{xcolor}

\usepackage{graphicx}
\usepackage{subcaption}
\usepackage{bm}
\usepackage{booktabs}
\usepackage{siunitx}
\usepackage{colortbl, color}

\usepackage{lipsum}
\usepackage{comment}
\usepackage{svg}
\usepackage{tikz}
\usetikzlibrary{positioning, arrows.meta, chains, calc}

\definecolor{cvprblue}{rgb}{0.21,0.49,0.74}
\definecolor{lblue}{rgb}{0.9,0.95,1}
\definecolor{lpurple}{rgb}{0.35,0.25,0.55}
\definecolor{lgreen}{rgb}{0.95,1,0.95}
\definecolor{sblue}{rgb}{0,0.45,1}
\definecolor{lgray}{gray}{0.95}
\definecolor{lyellow}{rgb}{1,1,0.92}

\usepackage[pagebackref,breaklinks,colorlinks,citecolor=cvprblue]{hyperref}

\def\paperID{*****} % *** Enter the Paper ID here
\def\confName{CVPR}
\def\confYear{2024}

\title{Deep RAW Image Super-Resolution. A NTIRE 2024 Challenge Survey}

\author{
Marcos V. Conde~$^{*\dagger}$ \and
Florin-Alexandru Vasluianu~$^{*}$ \and
Radu Timofte~$^{*}$ \and
Jianxing Zhang \and
Jia Li \and
Fan Wang \and
Xiaopeng Li \and
Zikun Liu \and
Hyunhee Park \and
Sejun Song \and
Changho Kim \and
Zhijuan Huang \and
Hongyuan Yu \and
Cheng Wan \and
Wending Xiang \and
Jiamin Lin \and
Hang Zhong \and
Qiaosong Zhang \and
Yue Sun \and
Xuanwu Yin \and
Kunlong Zuo \and
Senyan Xu \and
Siyuan Jiang \and
Zhijing Sun \and
Jiaying Zhu \and
Liangyan Li \and
Ke Chen \and
Yunzhe Li \and
Yimo Ning \and
Guanhua Zhao \and
Jun Chen \and
Jinyang Yu \and
Kele Xu \and
Qisheng Xu \and
Yong Dou \and
}

\begin{document}

\maketitle

\let\thefootnote\relax\footnotetext{$*$ Marcos V. Conde ($\dagger$ corresponding author, project lead), Florin-Alexandru Vasluianu, amd Radu Timofte are the challenge organizers, while the other authors participated in the challenge and survey. \\
$*$~University of W\"urzburg, CAIDAS \& IFI, Computer Vision Lab.\\ 
NTIRE 2024 webpage:~\url{https://cvlai.net/ntire/2024}.\\
Code:~\url{https://github.com/mv-lab/AISP}} 

\begin{abstract}
This paper reviews the NTIRE 2024 RAW Image Super-Resolution Challenge, highlighting the proposed solutions and results. New methods for RAW Super-Resolution could be essential in modern Image Signal Processing (ISP) pipelines, however, this problem is not as explored as in the RGB domain. Th goal of this challenge is to upscale RAW Bayer images by 2x, considering unknown degradations such as noise and blur. In the challenge, a total of 230 participants registered, and 45 submitted results during thee challenge period. The performance of the top-5 submissions is reviewed and provided here as a gauge for the current state-of-the-art in RAW Image Super-Resolution.
\end{abstract}

%% narrow the gap between equations and sentences
\setlength{\abovedisplayskip}{1pt}
\setlength{\belowdisplayskip}{1pt}

\section{Introduction}

RAW Image Super-Resolution represents an active research direction, aiming at upscaling hardware-specific RAW image representations, while dealing with hardware characteristic properties, often depending on the technical  implementation of the camera product. The lack of standardization in camera Image Processing Signal (ISP) implementation induces plenty variety into the camera market segment, with plenty corrections made through image processing algorithms \cite{xu2019rawsr, conde2024bsraw, conde2023perceptual, ignatov2021learnednpu, ignatov2020replacing}, overcoming the hardware limitations of various devices.

The RAW image represents the discretized and quantized representation of the image signal. Of course, both the aforementioned operation depend on the sensor nature, with different sensor types, different spatial resolutions or color resolutions (the number of bits used for quantization). Naturally, portable camera devices, that are usually subject to extreme limitations in terms of size, power supply, and used optics are constricted in terms of achieved image quality, usually employing low color resolutions and spatial resolutions. However, these specific hardware implementations are currently leading the market in terms of new acquired devices, with cameras becoming ubiquitous and universal accessible. Therefore, the image information space is also characterized by a high availability of images corresponding to these devices, with complex ISP systems  ~\cite{conde2022model,heide2014flexisp,schwartz2018deepisp,zhang2019zoom} mapping the RAW image representation to a perceptually meaningful RGB counterpart. 

RAW image upscaling remains also relevant in the professional photography field. Given the strong development of Internet-based image transfer services, the drive for high resolution images is now higher than ever before. However, building a professional photography or videography setup is a difficult challenge, since the market is divided by various application-specific parameters, with the acquisition costs remaining the main limiting factor for a new investment. Thus, it is important to develop algorithms which are robust to the various hardware limitations affecting the low-cost amateur photography systems, matching the characteristics of the high resolution sensors used in highly professional applications. 

%%%%%%%%%%%%%%%%%%%%%%%%%%%%%%%%%%%%
\begin{figure*}[t]
    \centering
    \setlength{\tabcolsep}{1pt}
    \begin{tabular}{c c c c}
         \includegraphics[width=0.245\linewidth]{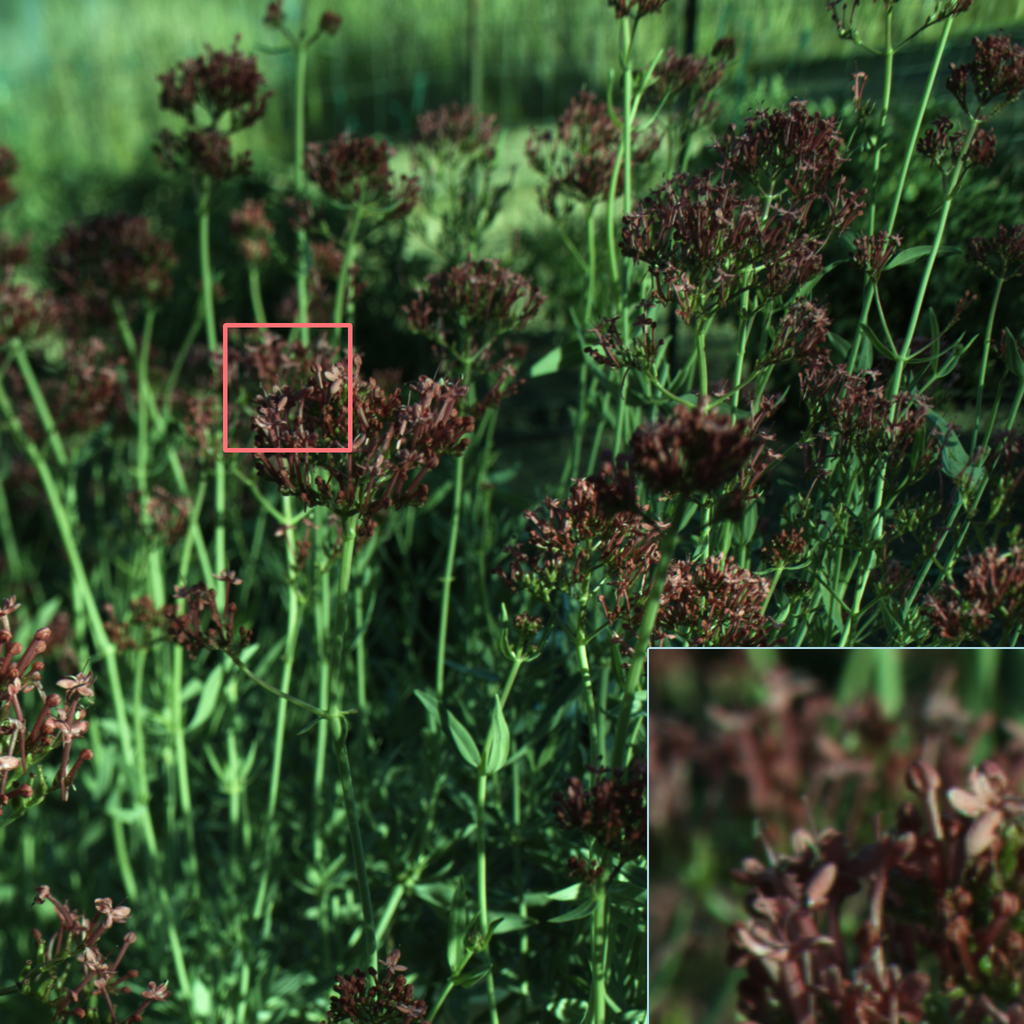} & 
         \includegraphics[width=0.245\linewidth]{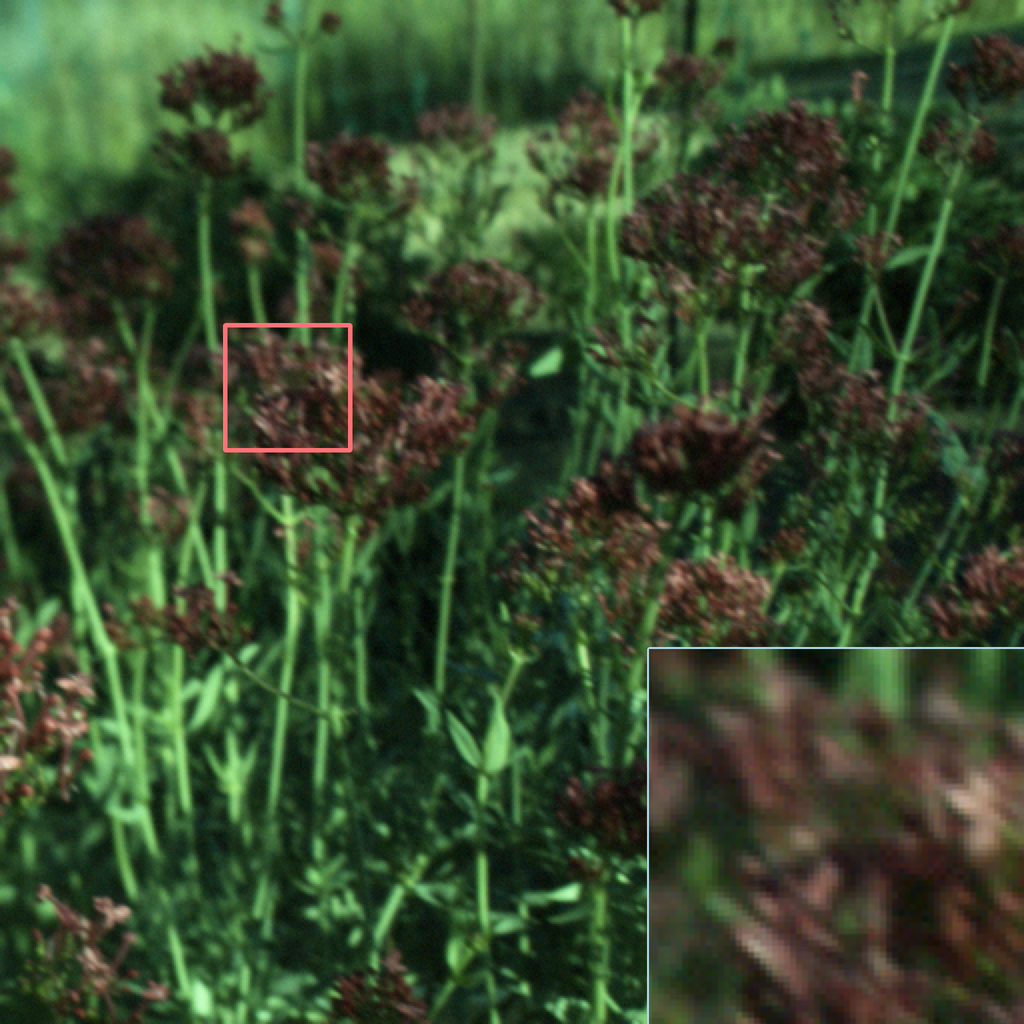} & 
         \includegraphics[width=0.245\linewidth]{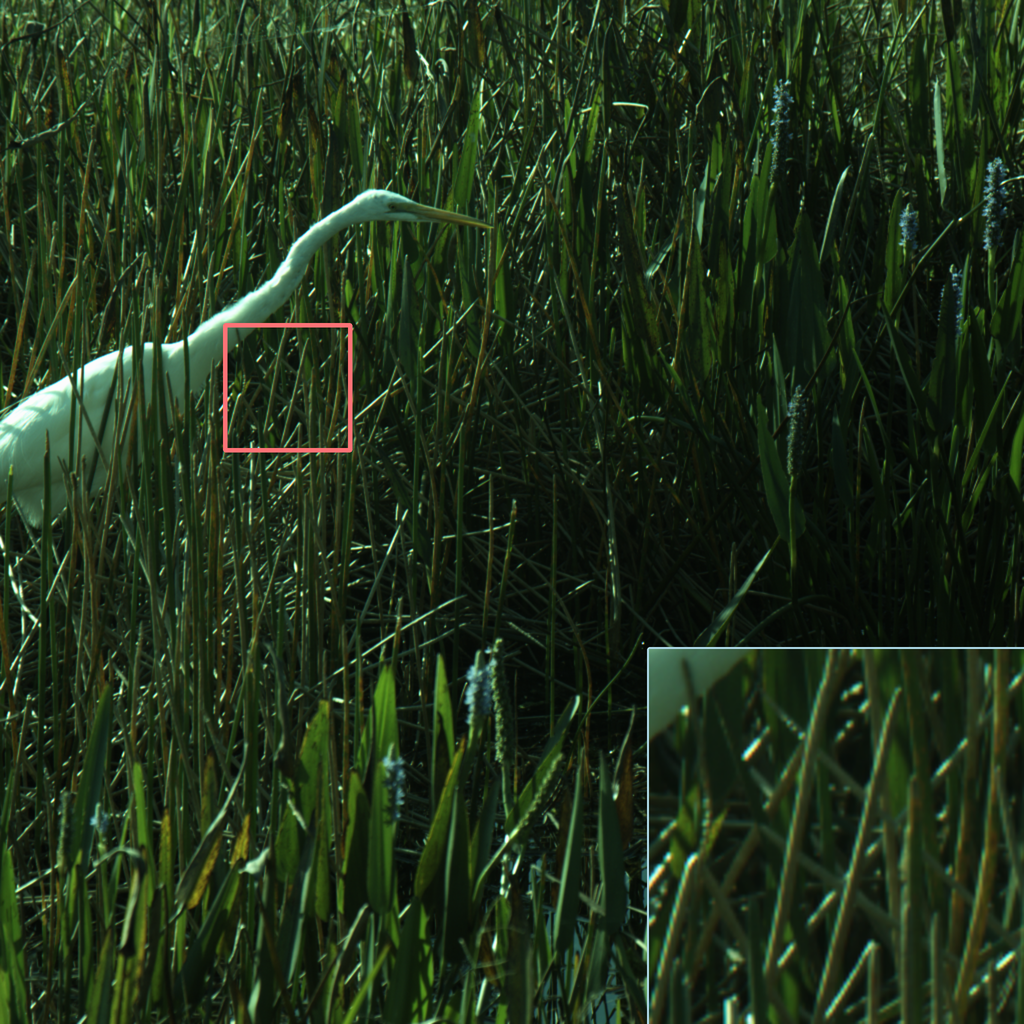} & 
         \includegraphics[width=0.245\linewidth]{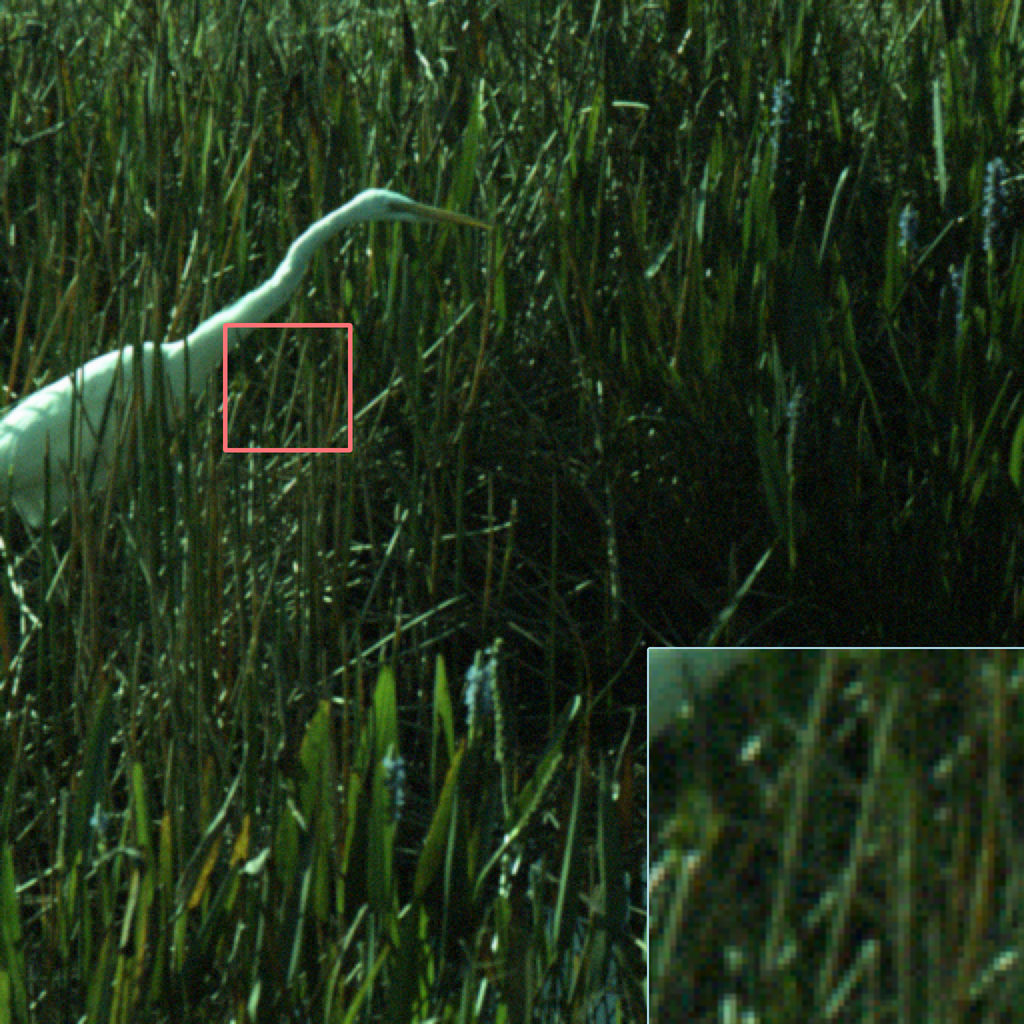}
         \tabularnewline
         \includegraphics[width=0.245\linewidth]{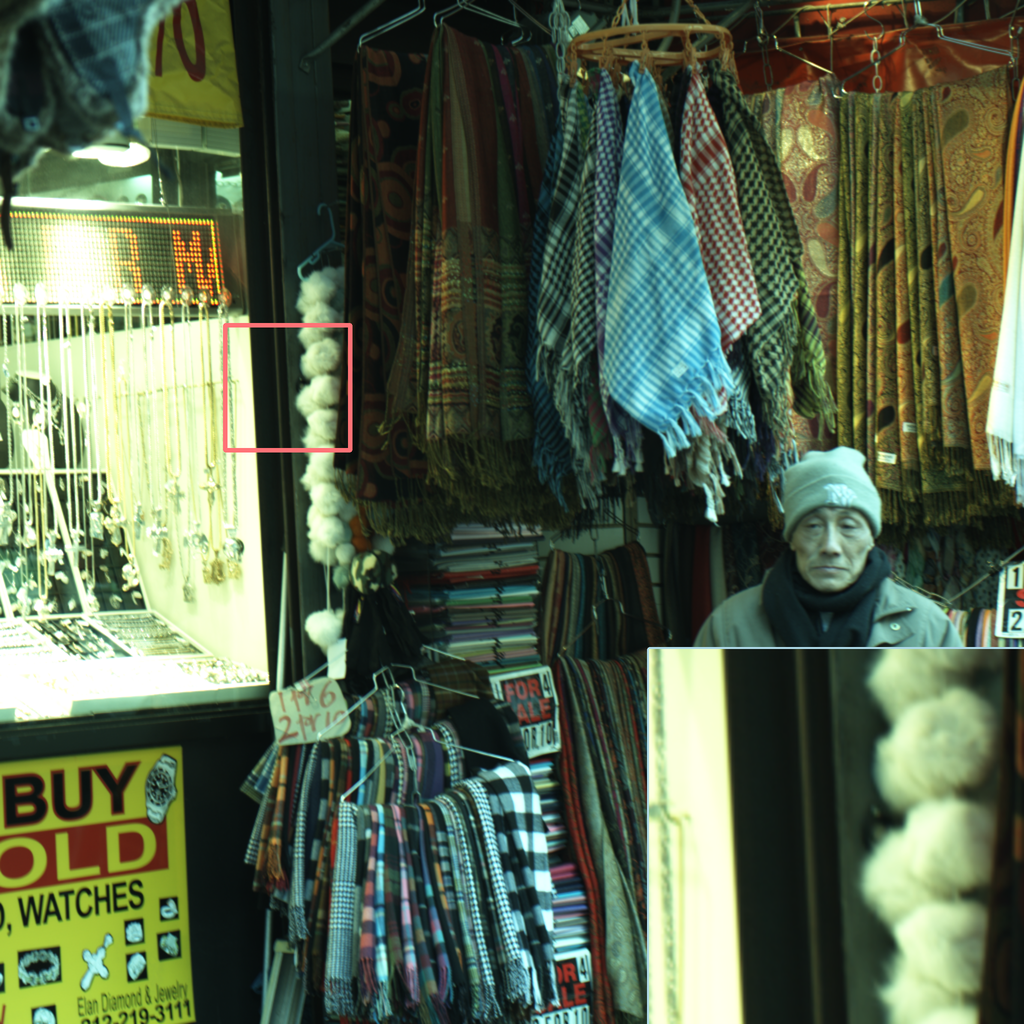} & 
         \includegraphics[width=0.245\linewidth]{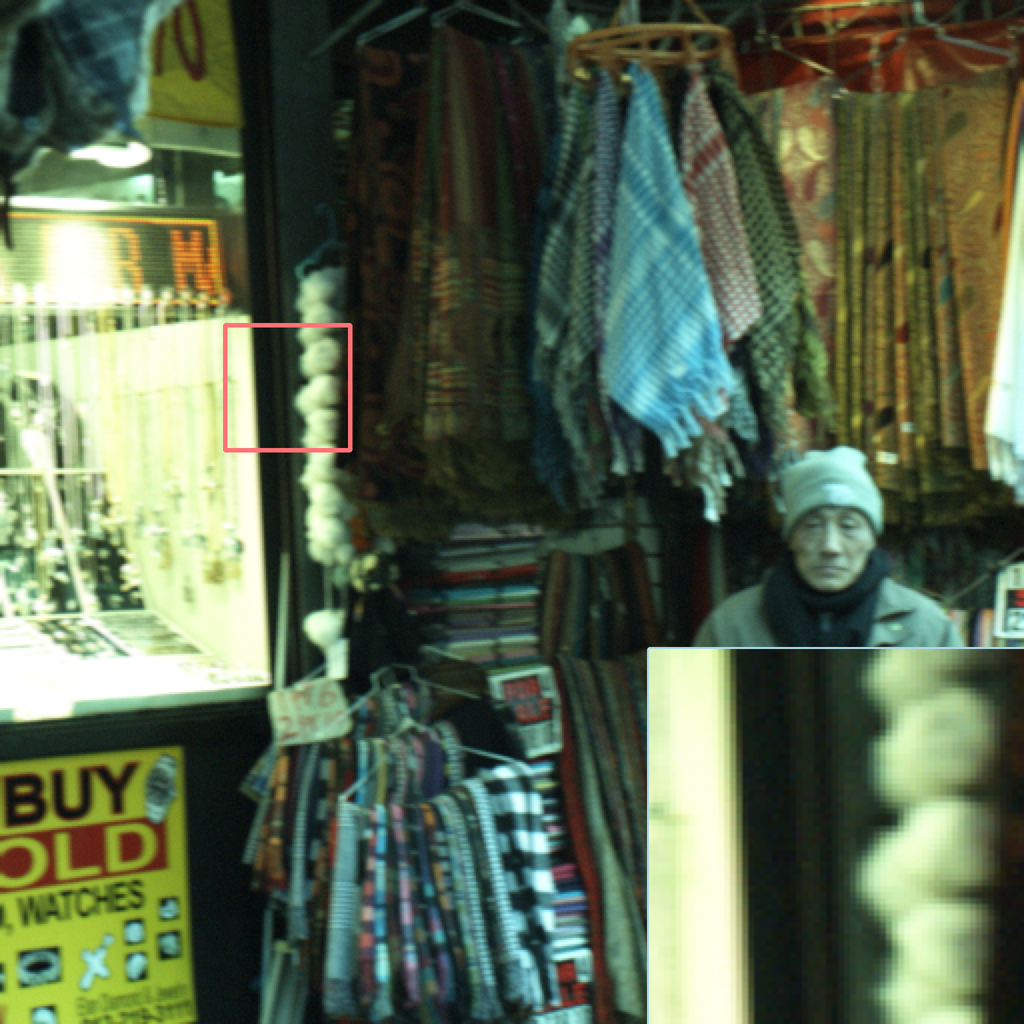} & 
         \includegraphics[width=0.245\linewidth]{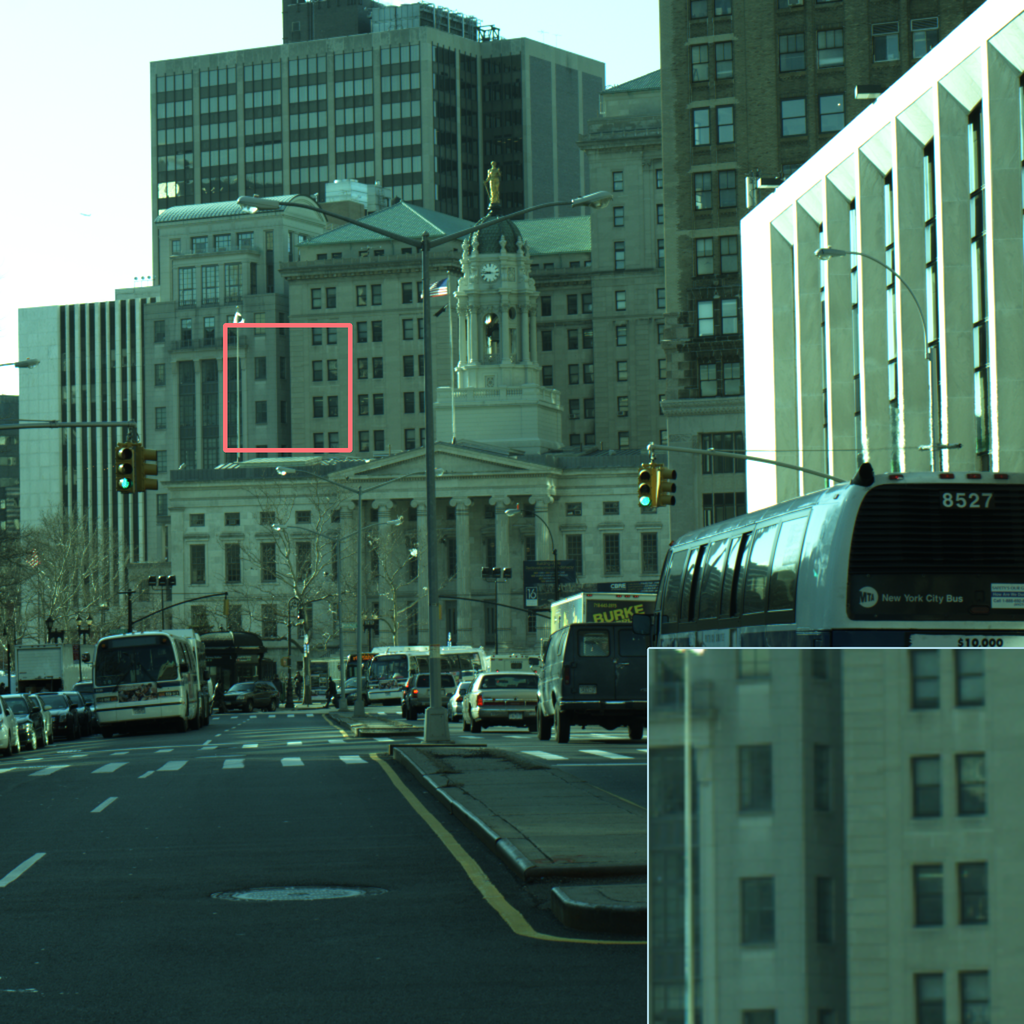} & 
         \includegraphics[width=0.245\linewidth]{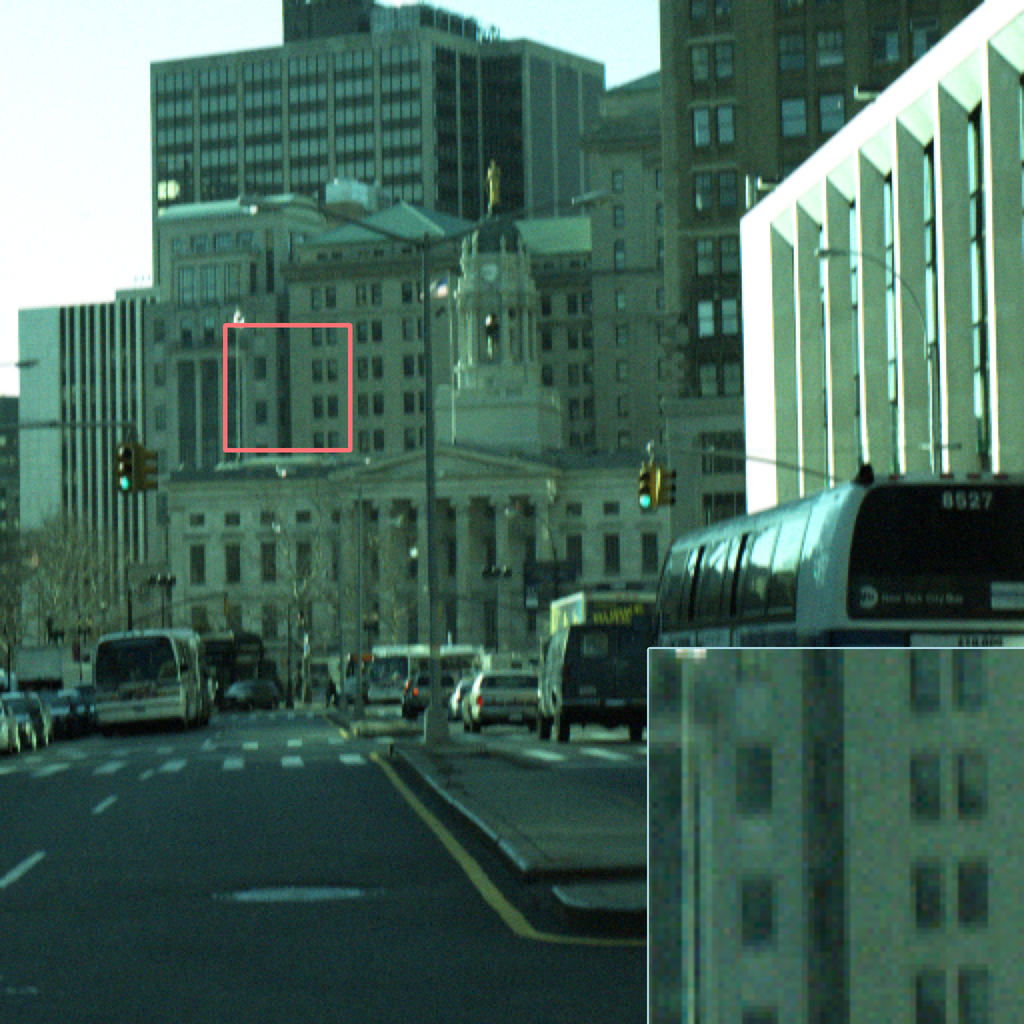}
         \tabularnewline
         HR Ground-truth & LR Input & HR Ground-truth & LR Input 
    \end{tabular}
    \caption{Samples of the \textbf{NTIRE 2024 RAW Image Super-Resolution Challenge} testing set.}
    \label{fig:test-samples}
    %kind of teaser image
    \end{figure*}
%%%%%%%%%%%%%%%%%%%%%%%%%%%%%%%%%%%%

RAW information is extremely important, since the RAW data directly correlates with scene radiance, with the discretization and quantization being the only non linear operations affecting the naturally continuous radiance signal measure in a photography. This property of the RAW data is extremely beneficial in analyzing typical image degradations as acquisition noise or image blur.
Futher away from the sensor representation, ISP represents a sequence of highly non-linear operations, happening with irrevocable loss of information ~\cite{brooks2019unprocessing, karaimer2016software}. This further complicates the image restoration task, with complex algorithms needing every achievable bit of signal based variance ~\cite{brooks2019unprocessing, xu2019rawsr}.
Considering all the aforementioned factors, RAW image processing poses significant advantages over the standard sRGB representation, with superior performance in a multitude of low-level imagery applications like image denoising~\cite{abdelhamed2018high, brooks2019unprocessing, mildenhall2018burst}, deblurring~\cite{conde2024bsraw}, exposure adjustment~\cite{hasinoff2016burst}, and image super-resolution~\cite{xu2019rawsr, zhang2019zoom, qian2019trinity, yue2022real, conde2024bsraw}.

%%%% INTRO 2 - Image Restoration and SR

% Nevertheless, only a small fraction of low-vision research directly engages with RAW data, mainly due to the greater abundance and accessibility of general-purpose sRGB images. % - this is crucial for the effective training of deep neural networks (DNNs).

Due to the lack of standardization at the hardware implementation level, RAW images are characterized by vendor or even product specific properties, which are explainable only with private, with the photography products shielded by implementation patents or trade secrets. Coupled to the abundance of the standard sRGB representation, most of the existing high complexity image restoration algorithms \cite{zamir2022restormer, liang2021swinir, conde2023perceptual, chen2022simple} are specifically designed for compressed or uncompressed RGB image or video. 

As a sub-task of Image Restoration, cutting-edge Single Image Super-Resolution (SISR) algorithms~\cite{liang2021swinir, conde2022swin2sr, agustsson2017ntire} follow the same data specific, even if they are relying on deep convolutional networks or Transformer~\cite{zamir2022restormer, liang2021swinir} architectures. 
One of the largest drawbacks characterizing these algorithms is represented by them being limited  by the quality of the data used for optimization. Various image restoration applications are characterized by extreme difficulties in acquiring real domain paired data \cite{abdelhamed2018high, li2019single}, driving the need for realistic and relevant data synthesis \cite{conde2024bsraw, xu2019rawsr}.  Accurately modeling application specific degradations in the sRGB representation proves extremely difficult, given the fact that the highly nonlinear ISP characteristic shifts also the physical characteristics of the degradation appearance. This represents the main factor limiting the performance of these algorithms in real applications deployment, with the gap observable at the data level being difficult to overcome with model-free algorithms.

Therefore, studying RAW data becomes a crucial step in driving the general image restoration performance improvement trend. Consequently, RAW Image Super-Resolution applications can benefit from the increased variance signal, with algorithms robust to fine architectural properties of the involved devices. Developing highly effective algorithms like the ones well-established in the sRGB domain \cite{conde2023perceptual, shi2016subpixel, 9897883} can prove a step ahead in the development of high performance imagery applications, with cost effective devices. 

Thus, in this work we are presenting the solutions submitted for the NTIRE 2024 RAW Image Super Resolution Challenge. We are providing information regarding the challenge setup, with the task description and the challenge data properties characterizing the challenge dataset splits.  We are also listing information regarding the challenge participants, with their teams and affiliations. 

In \cref{sec:challenge} we describe the challenge dataset and evaluation, and we discuss the overall results. 
In \cref{sec:teams} we provide detailed descriptions of the best solutions.

%%%%%%%%%%%%%%%%%%%%%%%%%%%%%%%%%%%%%%%%%5
%%%% BENCHMARK START PAGE 2

\begin{table*}[!ht]
    \centering
    \resizebox{\linewidth}{!}{
    \begin{tabular}{l l c c c c c c c}
    \toprule
    \rowcolor{lgray}  Team & Method & Validation 1MP & Test 1MP & Test 12MP & \# Params. (M) & Train E2E & Train Res. \\
    \toprule

    Samsung    & 2-Stage w/ FPL & 43.40 / 0.99 & 43.443 / 0.986 & 43.858 / 0.988 & 53.7 & Yes & 384 \\
    
    XiaomiMMAI & EffectiveSR & 43.38 / 0.99 & 43.249 / 0.986 & NA & 20.9 & No & 64 \\

    USTCX & RBSFormer~\cite{jiang2024rbsformer} & 43.21 / 0.99  & 42.493 / 0.984 & 43.649 / 0.987 & 3.3 & Yes & 112 \\

    McMaster & SwinFSR Raw & 42.48 / 0.98 & 42.366 / 0.984 & NA & 6.64 & Yes & 256 \\

    \rowcolor{lblue} & BSRAW~\cite{conde2024bsraw} & 42.25 / 0.98  & 42.106 / 0.984 & 42.853 / 0.986 & 1.5 & Yes &  248 \\

    NUDT RSR    & SAFMN FFT & 41.81 / 0.98 & 41.621 / 0.982  &  NA & 0.27 & No & 128 - 448 \\

    \rowcolor{lblue} & Interpolation~\cite{conde2024bsraw} & 35.95 / 0.95 & 36.038 / 0.952 & 36.926 / 0.956 &  &  &  \\
    
    \bottomrule
    \end{tabular}
    }
    
    \caption{We provide \textbf{PSNR/SSIM} results on the validation set (40 images), the complete testing set (200 images), and the testing set at full-resolution (12MP) RAW images~\cite{conde2024ntire_raw}. All the fidelity metrics are calculated in the RAW domain. ``NA" indicates the results are not available for the method.
    We highlight two baseline methods. We also report the number of parameters of each method, if the method was trained end-to-end (Yes/No), and the image resolution used for training the models.
    }
    \label{tab:benchmark}
\end{table*}

\begin{comment}
\begin{table*}[]
    \centering
    \resizebox{\textwidth}{!}{
    \begin{tabular}{c|c|c|c|c|c|c}
        Input & Training Time & Train E2E & Extra Data & \# Params. (M) & GPU  \\
        \hline
         (128/256/352/448, 128/256/352/448, 4) & 3 days & No & No & 0.272 Million & RTX2080Ti \\
         (384,384,4) & 72h & Yes & No & 53.7 & A100 \\
         (64,64,4) & 80+h & No & No & 20.9 Million & A100 \\
         (112,112,3) & 9h & Yes & No & 3.3 Million & 4090 \\
         (256,256,4) & 96h & Yes & No & 6.64 Million & RTX4090 \\
    \end{tabular}
    }
    \caption{A brief view of our method.}
    \label{tab:my_label}
\end{table*}   
\end{comment}

%%%%%%%%%%%%%%%%%%%%%%%%%%%%%%%%%%%%%%%%%%%55

\paragraph{Related Computer Vision Challenges}
Our challenge is one of the NTIRE 2024 Workshop~\footnote{https://cvlai.net/ntire/2024/} associated challenges on: dense and non-homogeneous dehazing~\cite{ntire2024dehazing}, night photography rendering~\cite{ntire2024night}, blind compressed image enhancement~\cite{ntire2024compressed}, shadow removal~\cite{ntire2024shadow}, efficient super resolution~\cite{ntire2024efficientsr}, image super resolution ($\times$4)~\cite{ntire2024srx4}, light field image super-resolution~\cite{ntire2024lightfield}, stereo image super-resolution~\cite{ntire2024stereosr}, HR depth from images of specular and transparent surfaces~\cite{ntire2024depth}, bracketing image restoration and enhancement~\cite{ntire2024bracketing}, portrait quality assessment~\cite{ntire2024QA_portrait}, quality assessment for AI-generated content~\cite{ntire2024QA_AI}, restore any image model (RAIM) in the wild~\cite{ntire2024raim}, RAW image super-resolution~\cite{conde2024ntire_raw}, short-form UGC video quality assessment~\cite{ntire2024QA_UGC}, low light enhancement~\cite{ntire2024lowlight}.

\section{NTIRE 2024 RAWSR Challenge}
\label{sec:challenge}

\subsection{Dataset}

The challenge dataset is based on BSRAW~\cite{conde2024bsraw}. Following previous work~\cite{ xu2019rawsr, xu2020exploiting, conde2024bsraw}, we use images from the Adobe MIT5K dataset~\cite{fivek}, which includes images from multple Canon and Nikon DSLR cameras. 

The DSLR images are manually filtered to ensure diversity and natural properties (\ie remove extremely dark or overexposed images), we also remove the blurry images (\ie we only consider all-in-focus images).

The \textbf{pre-processing} is as follows: (i) we normalize all RAW images depending on their black level and bit-depth. (ii) we convert (``pack") the images into the well-known RGGB Bayer pattern (4-channels), which allows to apply the transformations and degradations without damaging the original color pattern information~\cite{liu2019learningrawaug}.

\noindent \textbf{Training:} We provide the participants 1064 $1024 \times 1024 \times 4$ clean high-resolution (HR) RAW images. The LR degraded images can be generated on-line during training using the degradation pipeline proposed in BSRAW~\cite{conde2024bsraw}. 

Such degradation pipeline considers different noise profiles, multiple blur kernels (PSFs) and a simple downsampling strategy to synthesize low-resolution (LR) RAW images. The participants can apply other augmentation techniques or expand the degradation pipeline to generate more realistic training data.

\subsection{Baselines}

We use BSRAW~\cite{conde2024bsraw} as the main baseline. The top performing challenge solutions improve the baseline performance, however, the neural networks are notably more complex in terms of design and computation.

\subsection{Results}

We use three testing splits: (i) Validation, 40 1024px images using during the model development phase. (ii) Test 1MP, 200 images of 1024px resolution. (iii) The same 200 test images at full-resolution $\approx12$MP. The participants process the corresponding LR RAW images (\eg $512 \times 512 \times 4$), and submit their results. Thus, the participants do not have access to the ground-truth images. 

We provide samples of the testing set in \cref{fig:test-samples}.

\begin{figure*}[t]
    \centering
    \setlength{\tabcolsep}{1pt}
    \begin{tabular}{c c c c}
         \includegraphics[width=0.245\linewidth]{figs/test-samples/in_190.png} & 
         \includegraphics[width=0.245\linewidth]{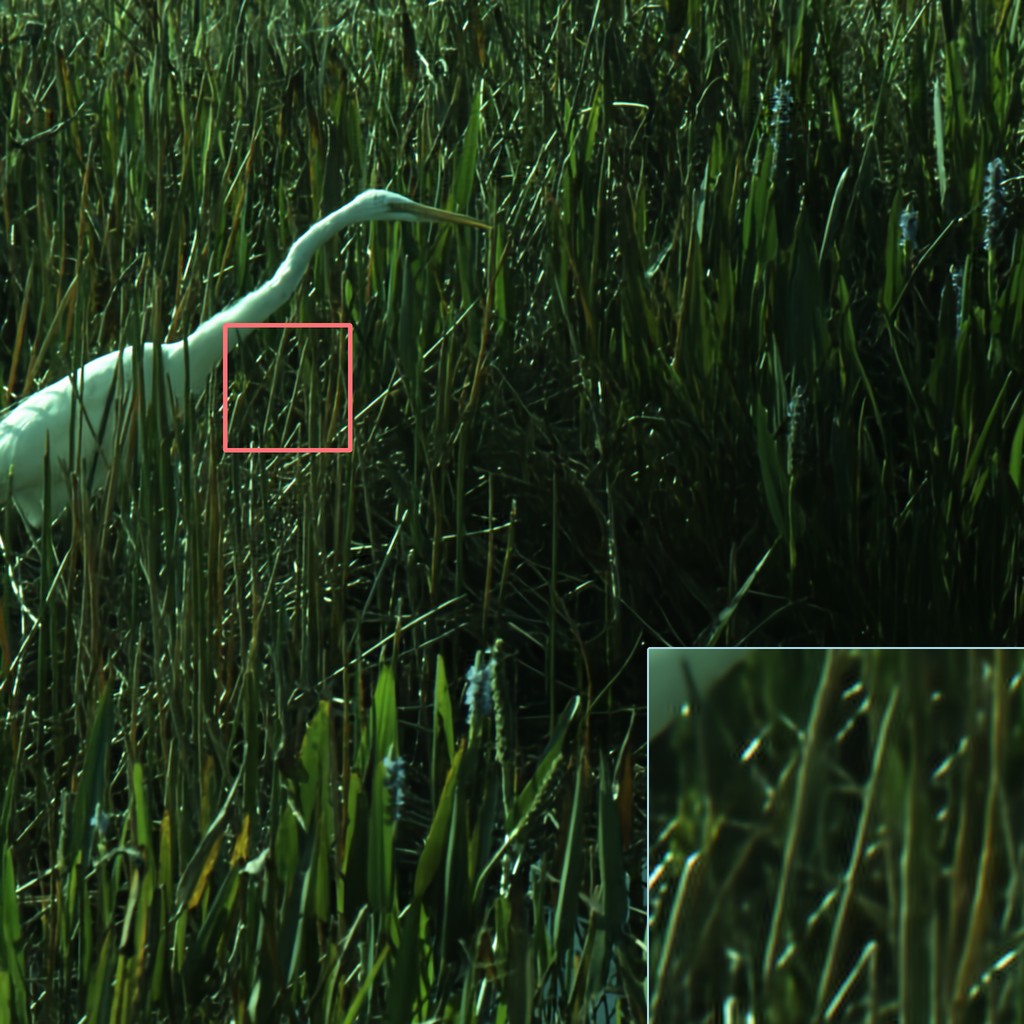} & 
         \includegraphics[width=0.245\linewidth]{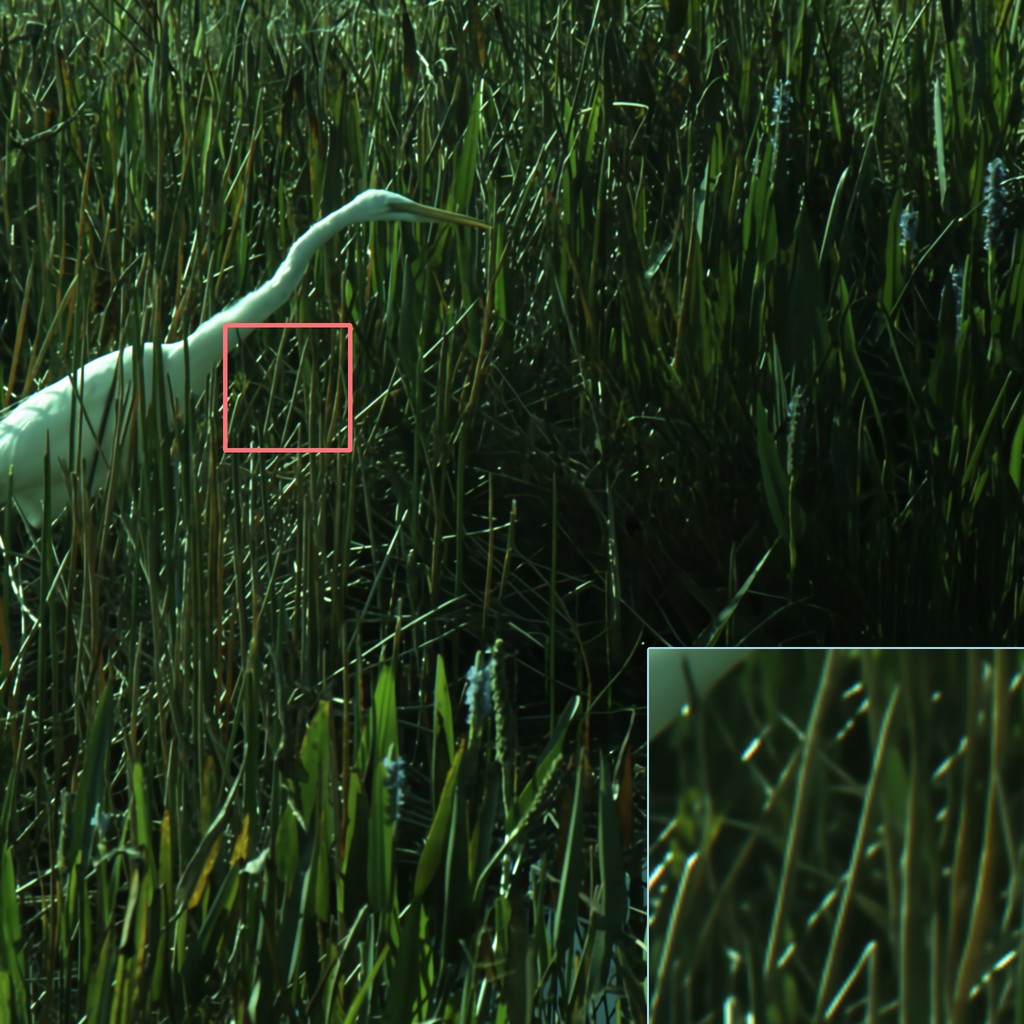} & 
         \includegraphics[width=0.245\linewidth]{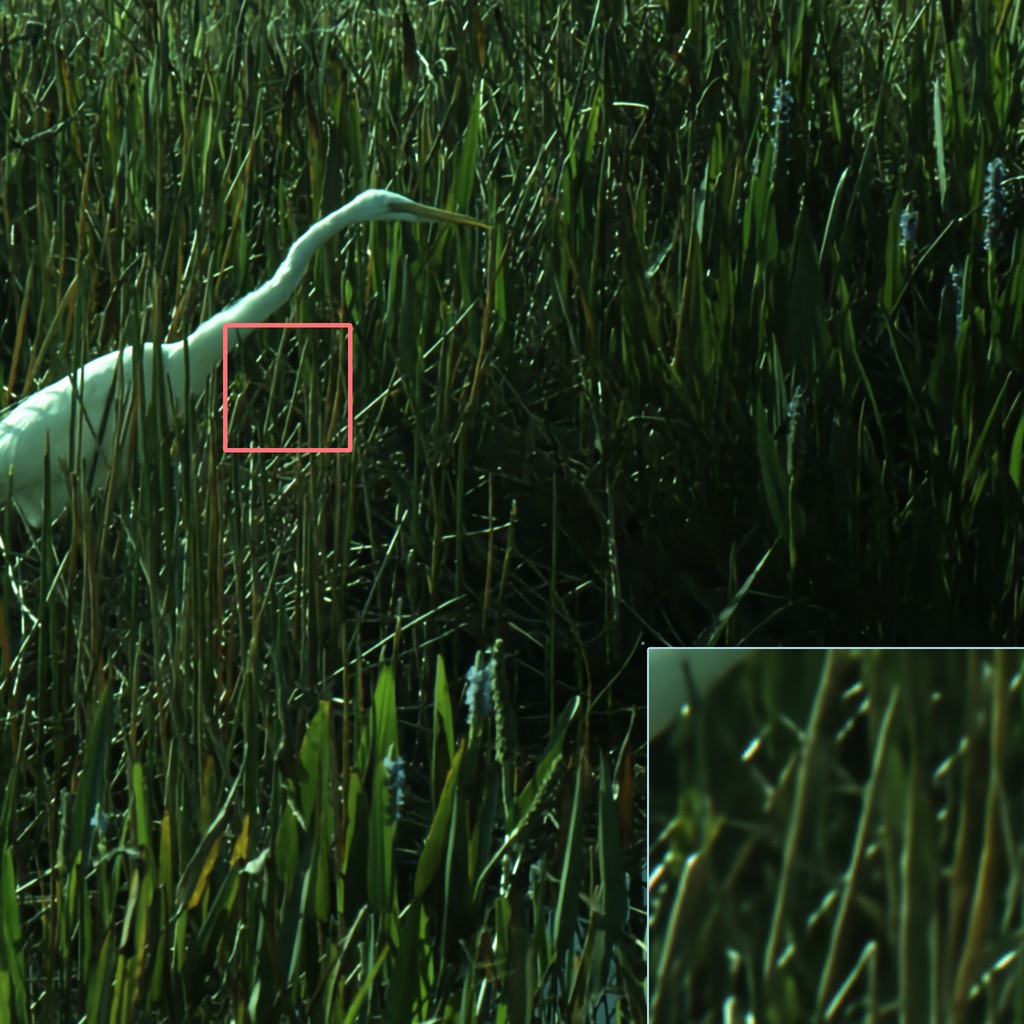}
         \tabularnewline
         LR Input & Team Samsung & EffectiveSR & RBSFormer~\cite{jiang2024rbsformer} \\
         
         \includegraphics[width=0.245\linewidth]{figs/test-samples/gt_190.png} & 
         \includegraphics[width=0.245\linewidth]{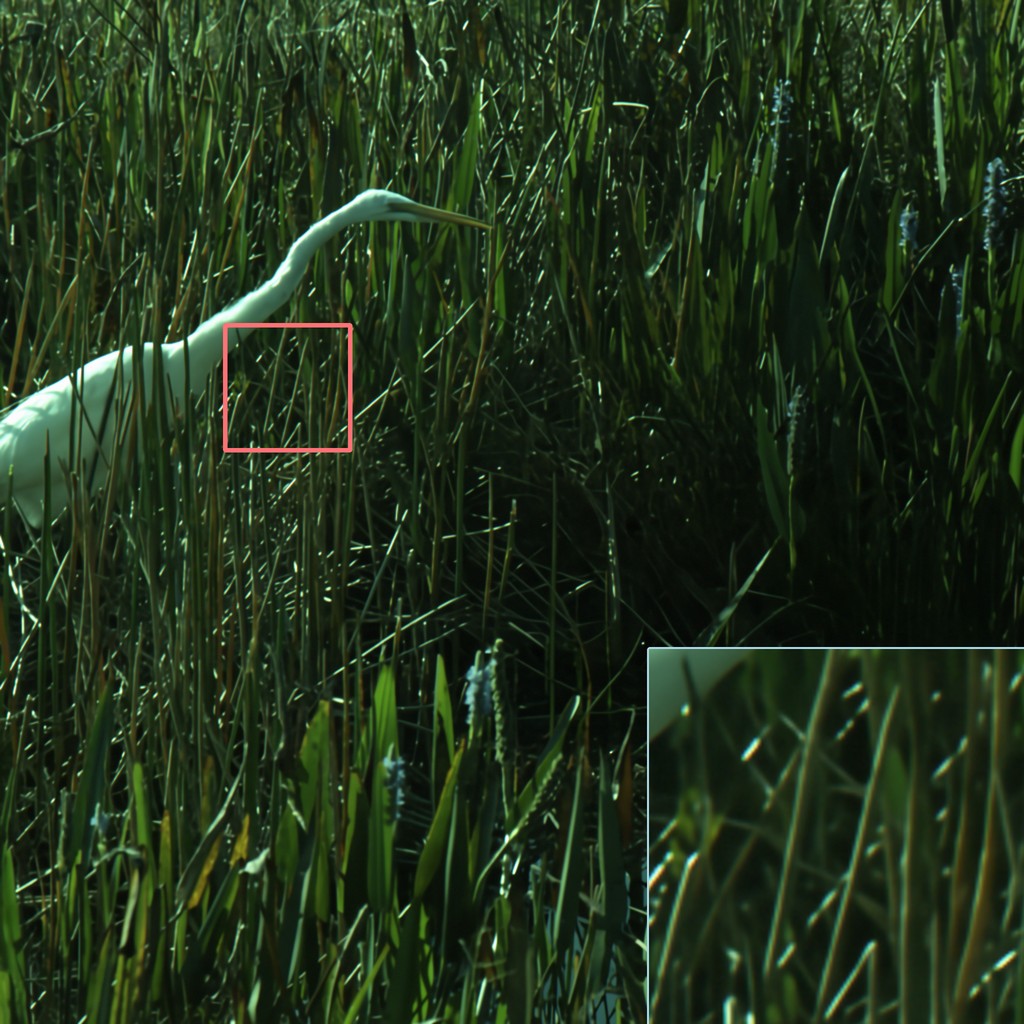} & 
         \includegraphics[width=0.245\linewidth]{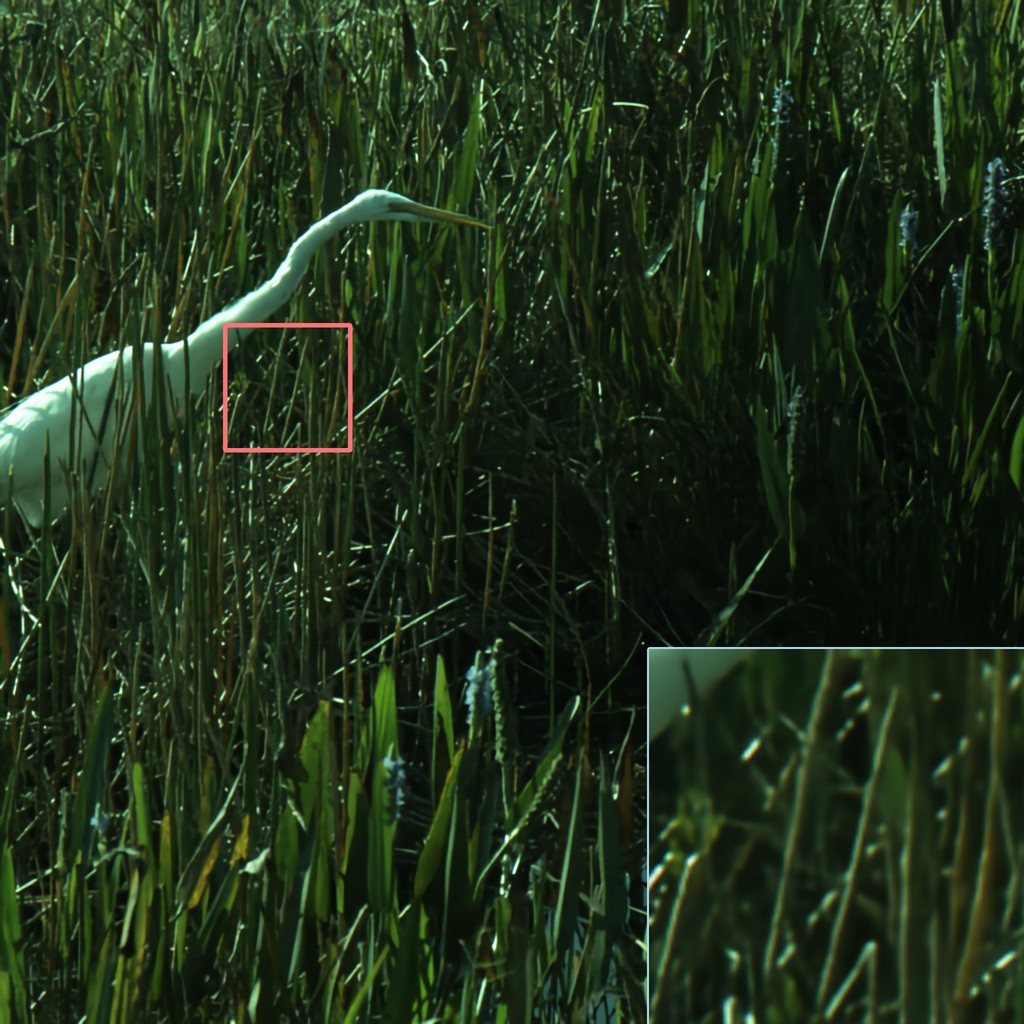} & 
         \includegraphics[width=0.245\linewidth]{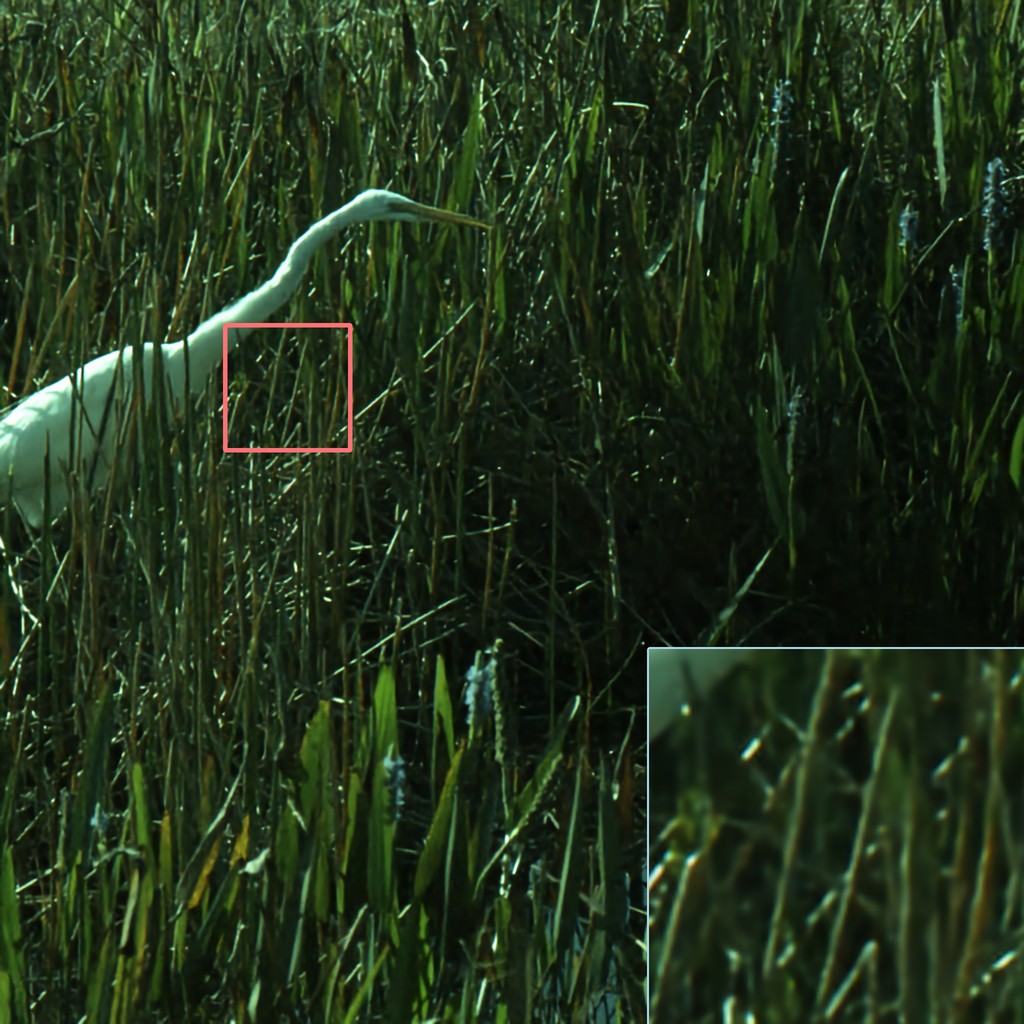}
         \tabularnewline
         HR Ground-truth & BSRAW~\cite{conde2024bsraw} & SwinFSR & SAFMN FFT \\
         % CROPS
         \includegraphics[width=0.245\linewidth]{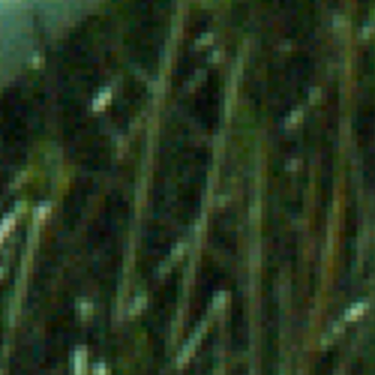} & 
         \includegraphics[width=0.245\linewidth]{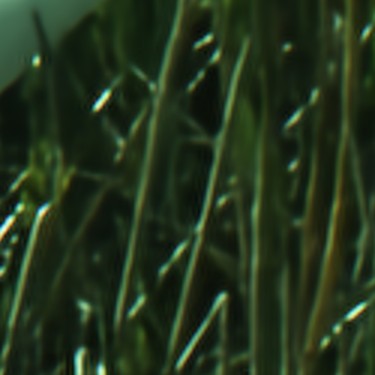} & 
         \includegraphics[width=0.245\linewidth]{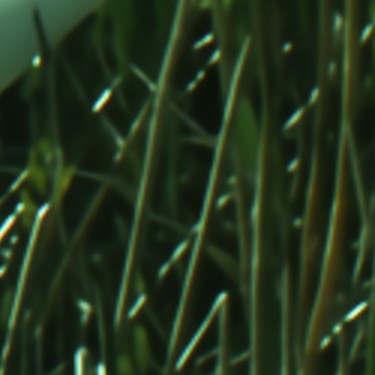} & 
         \includegraphics[width=0.245\linewidth]{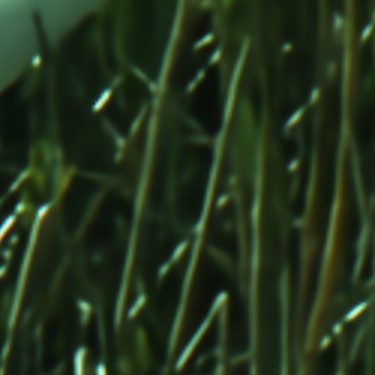}
         \tabularnewline
         LR Input & Team Samsung & EffectiveSR & RBSFormer~\cite{jiang2024rbsformer} \\
         
         \includegraphics[width=0.245\linewidth]{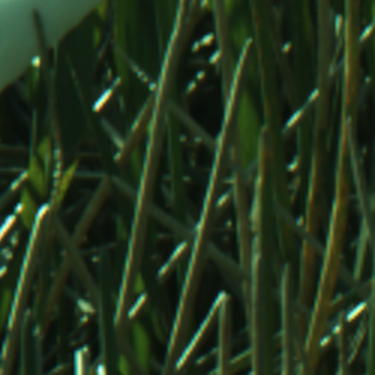} & 
         \includegraphics[width=0.245\linewidth]{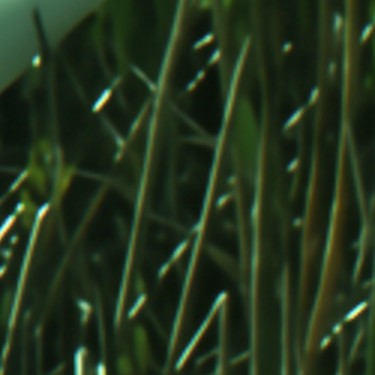} & 
         \includegraphics[width=0.245\linewidth]{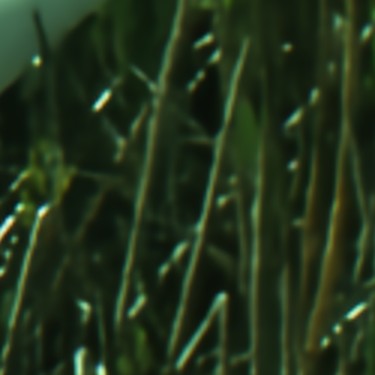} & 
         \includegraphics[width=0.245\linewidth]{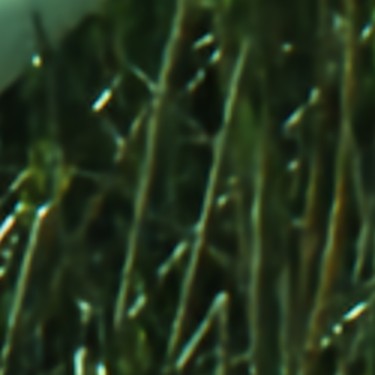}
         \tabularnewline
         HR Ground-truth & BSRAW~\cite{conde2024bsraw} & SwinFSR & SAFMN FFT
    \end{tabular}
    \caption{Visual comparison using the \textbf{NTIRE 2024 RAW Image Super-Resolution Challenge} testing set (\texttt{190.npz}). The HR resolution RAW images have $1024\times1024$ resolution and 4-channels (RGGB Bayer pattern). RAW images are visualized using bilinear demosaicing, gamma correction and tone mapping.}
    \label{fig:results1}
    \end{figure*}

\begin{figure*}[t]
    \centering
    \setlength{\tabcolsep}{1pt}
    \begin{tabular}{c c c c}
         \includegraphics[width=0.245\linewidth]{figs/test-samples/in_155.png} & 
         \includegraphics[width=0.245\linewidth]{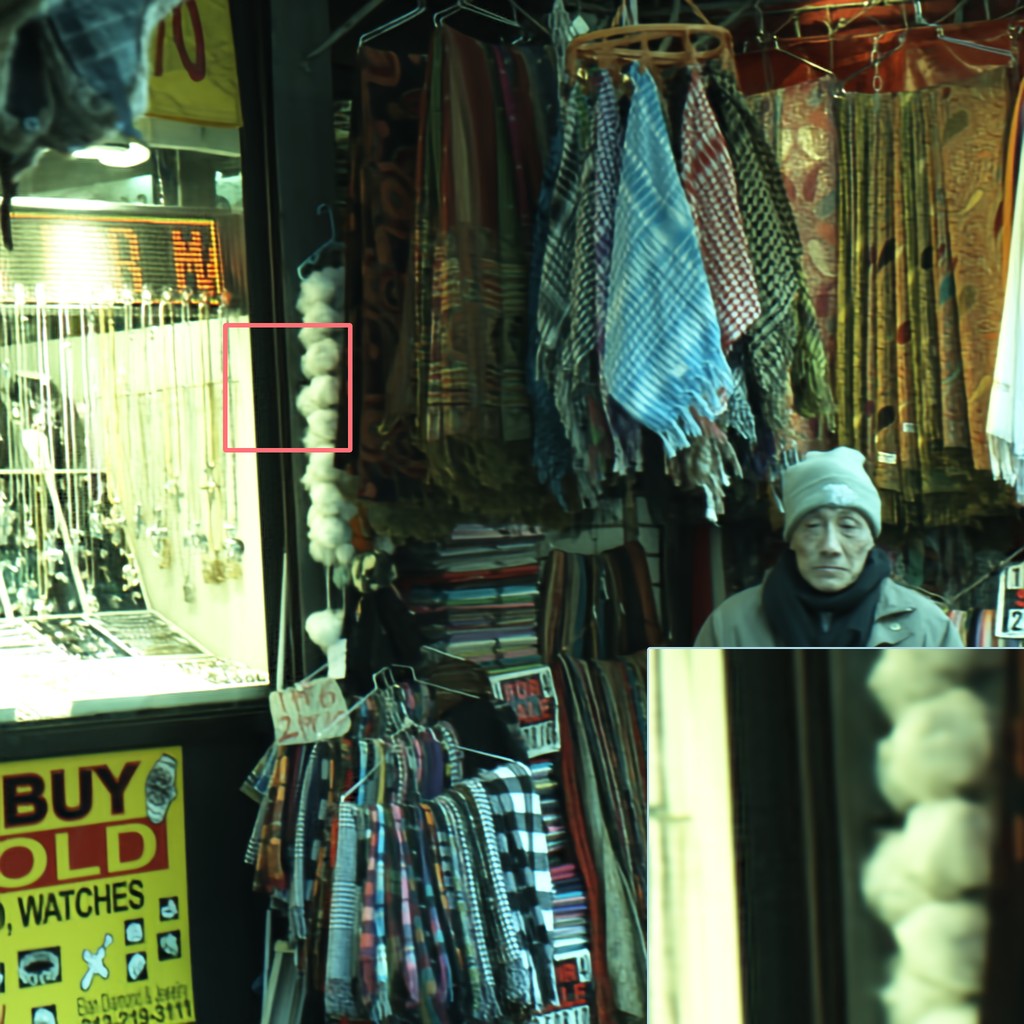} & 
         \includegraphics[width=0.245\linewidth]{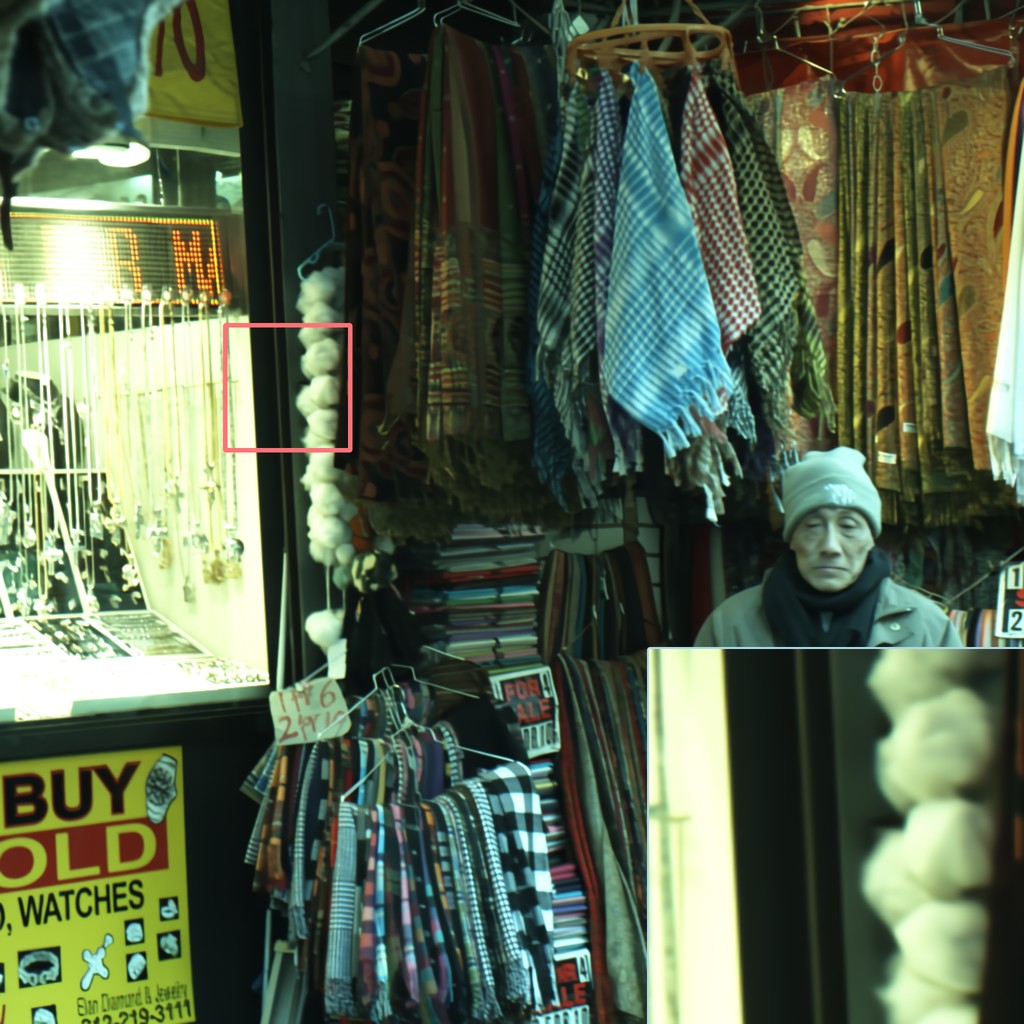} & 
         \includegraphics[width=0.245\linewidth]{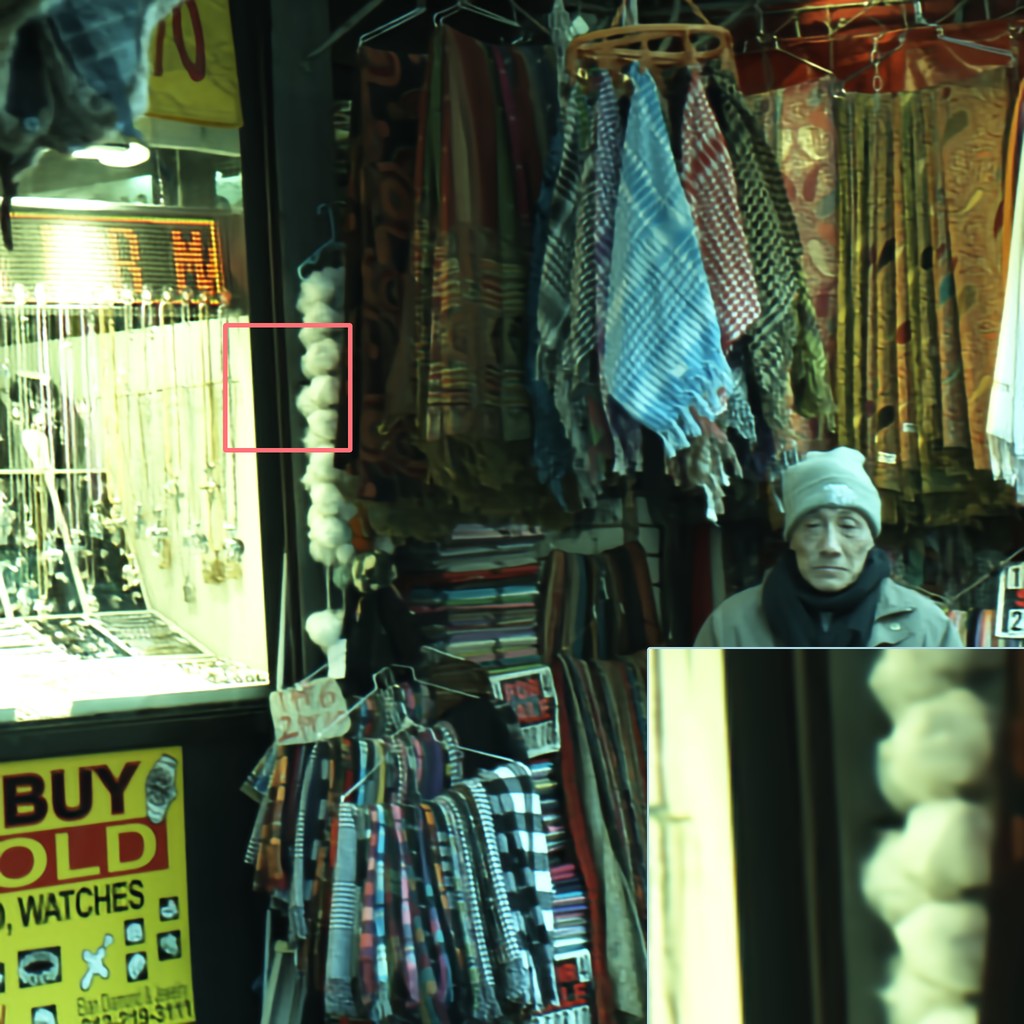}
         \tabularnewline
         LR Input & Team Samsung & EffectiveSR & RBSFormer~\cite{jiang2024rbsformer} \\
         
         \includegraphics[width=0.245\linewidth]{figs/test-samples/gt_155.png} & 
         \includegraphics[width=0.245\linewidth]{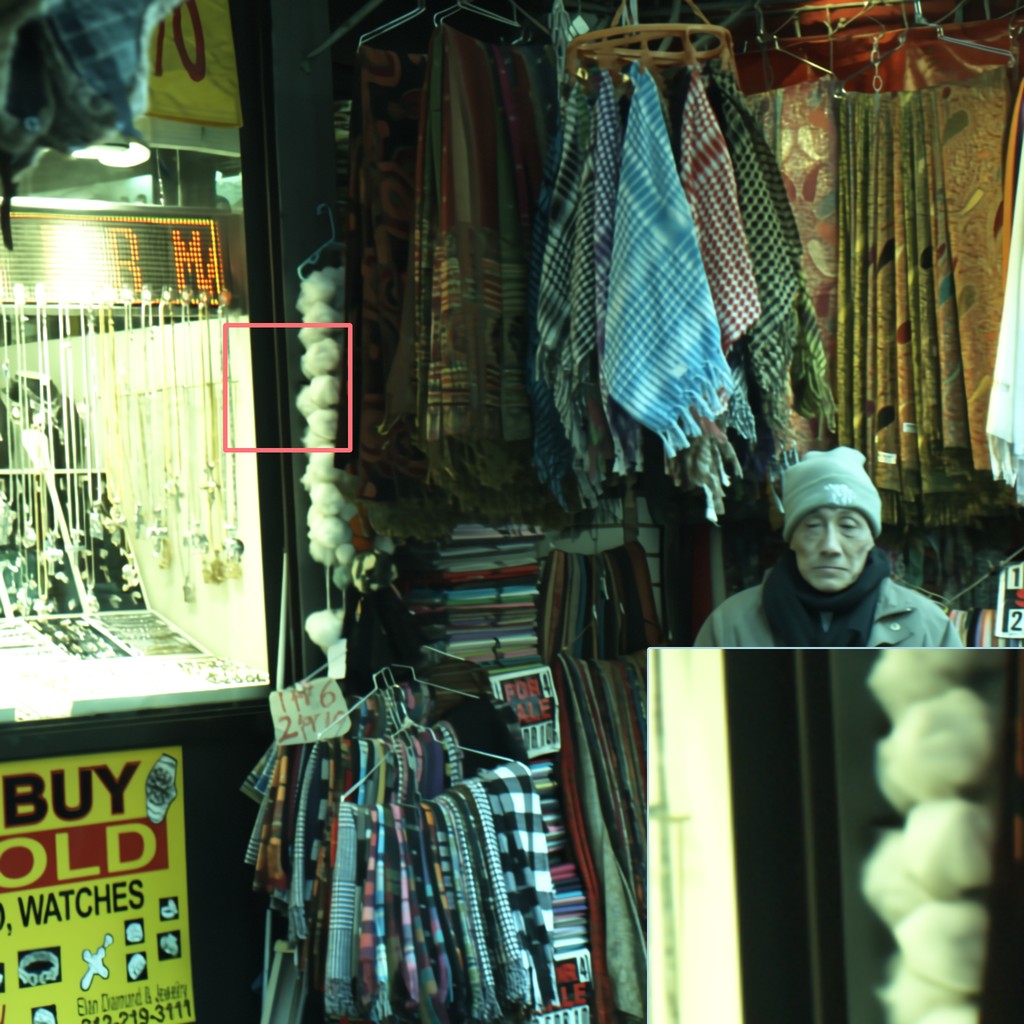} & 
         \includegraphics[width=0.245\linewidth]{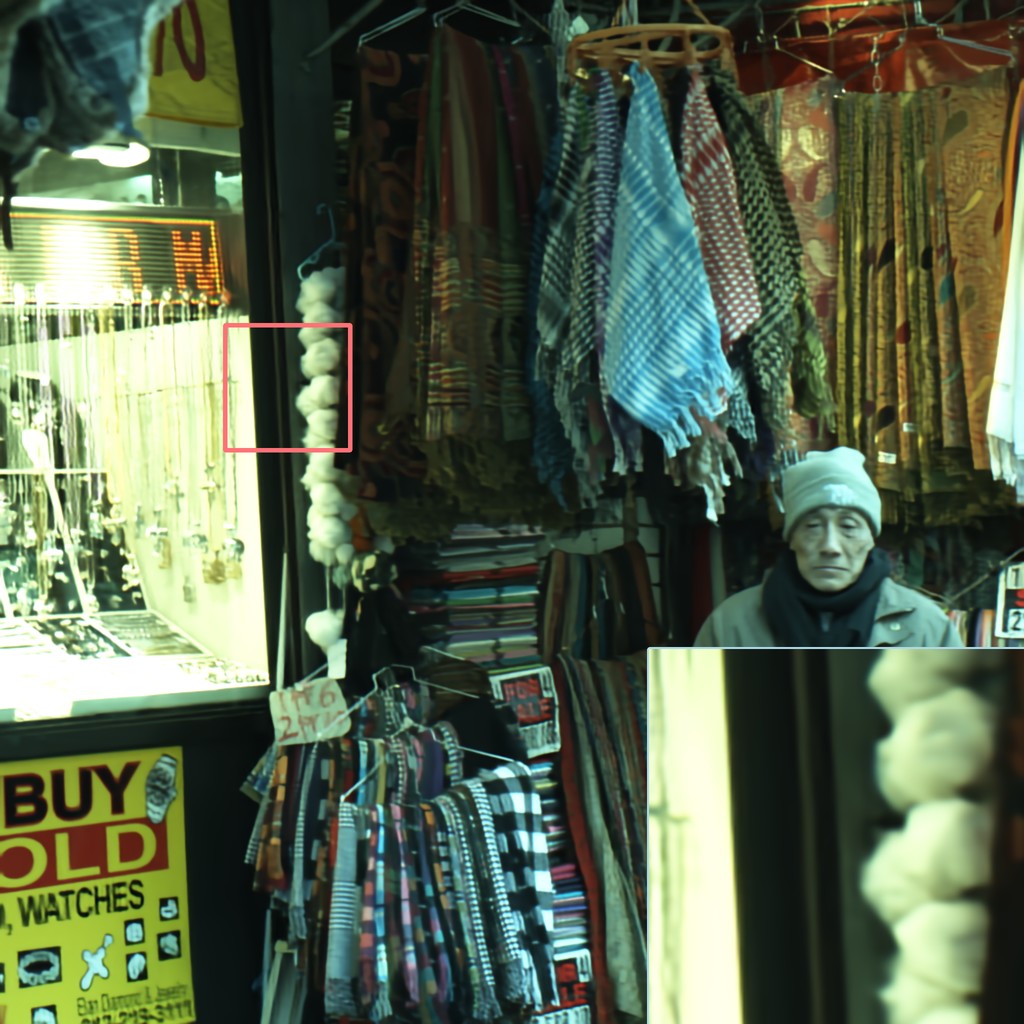} & 
         \includegraphics[width=0.245\linewidth]{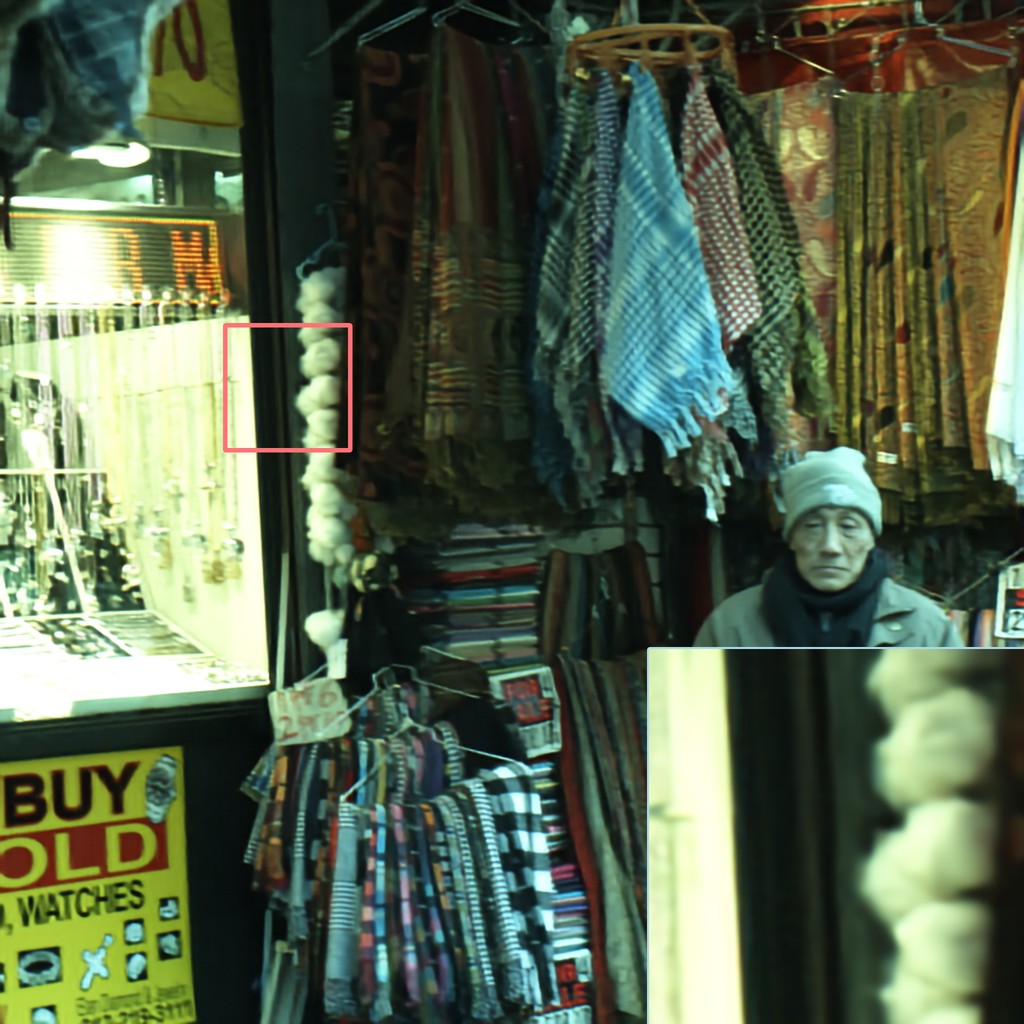}
         \tabularnewline
         HR Ground-truth & BSRAW~\cite{conde2024bsraw} & SwinFSR & SAFMN FFT \\
         % CROPS
         \includegraphics[width=0.245\linewidth]{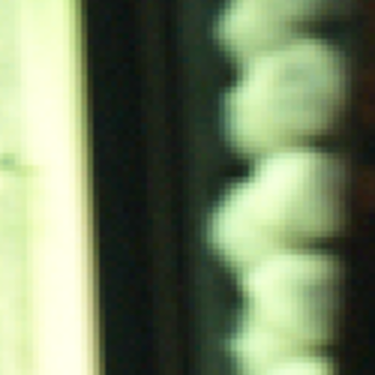} & 
         \includegraphics[width=0.245\linewidth]{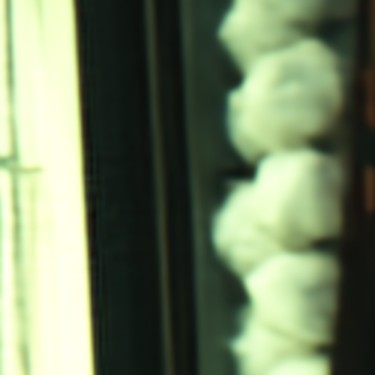} & 
         \includegraphics[width=0.245\linewidth]{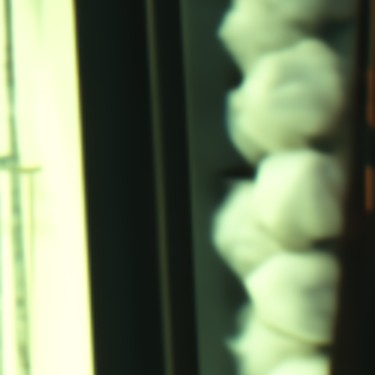} & 
         \includegraphics[width=0.245\linewidth]{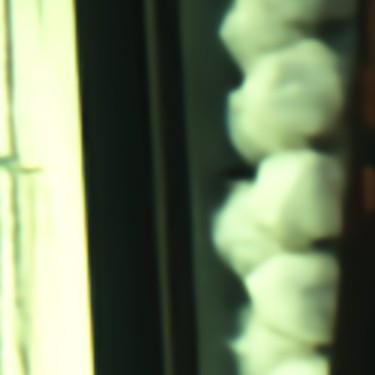}
         \tabularnewline
         LR Input & Team Samsung & EffectiveSR & RBSFormer~\cite{jiang2024rbsformer} \\
         
         \includegraphics[width=0.245\linewidth]{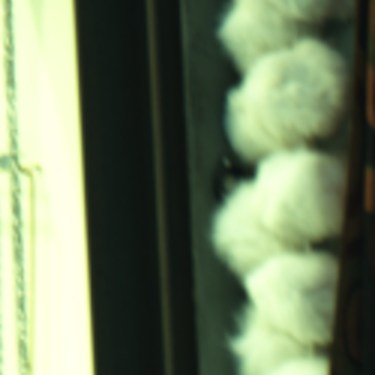} & 
         \includegraphics[width=0.245\linewidth]{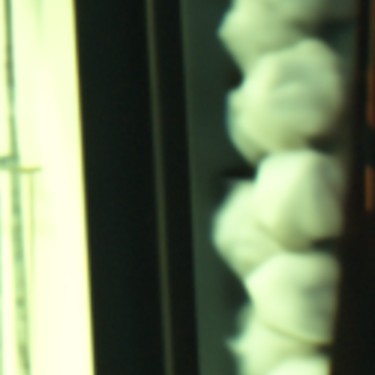} & 
         \includegraphics[width=0.245\linewidth]{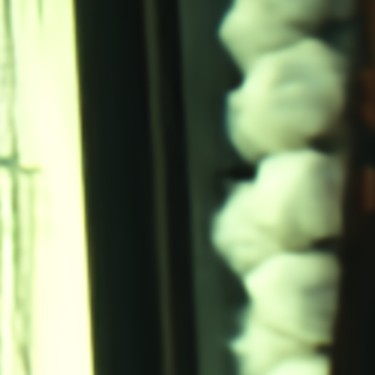} & 
         \includegraphics[width=0.245\linewidth]{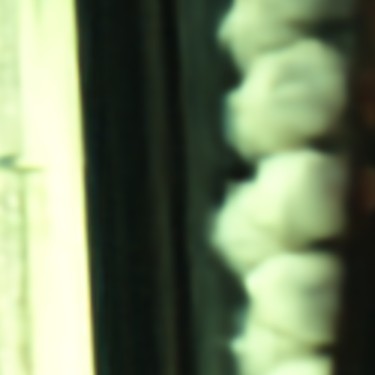}
         \tabularnewline
         HR Ground-truth & BSRAW~\cite{conde2024bsraw} & SwinFSR & SAFMN FFT
    \end{tabular}
    \caption{Visual comparison using the \textbf{NTIRE 2024 RAW Image Super-Resolution Challenge} testing set (\texttt{155.npz}). The HR resolution RAW images have $1024\times1024$ resolution and 4-channels (RGGB Bayer pattern). RAW images are visualized using bilinear demosaicing, gamma correction and tone mapping.}
    \label{fig:results2}
    \end{figure*}

\begin{figure*}[t]
    \centering
    \setlength{\tabcolsep}{1pt}
    \begin{tabular}{c c c c}
         \includegraphics[width=0.245\linewidth]{figs/test-samples/in_177.png} & 
         \includegraphics[width=0.245\linewidth]{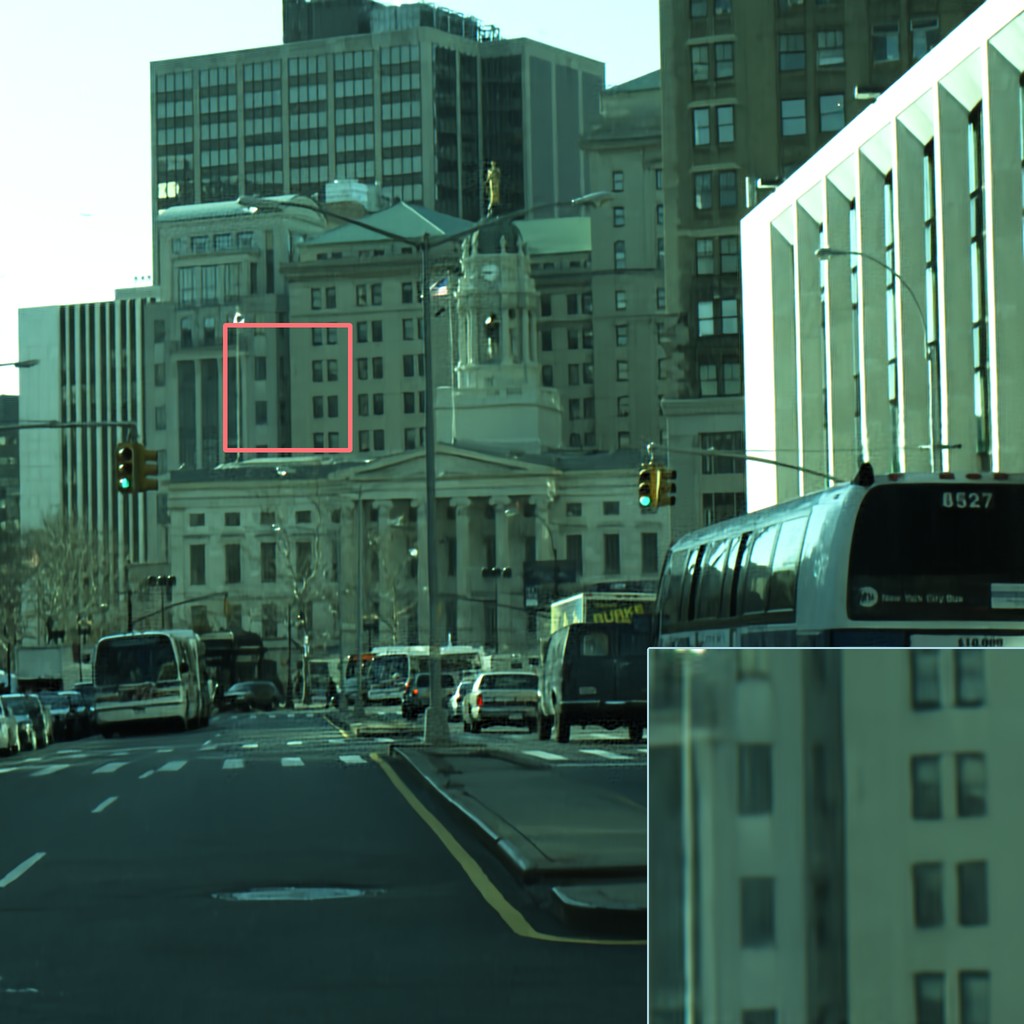} & 
         \includegraphics[width=0.245\linewidth]{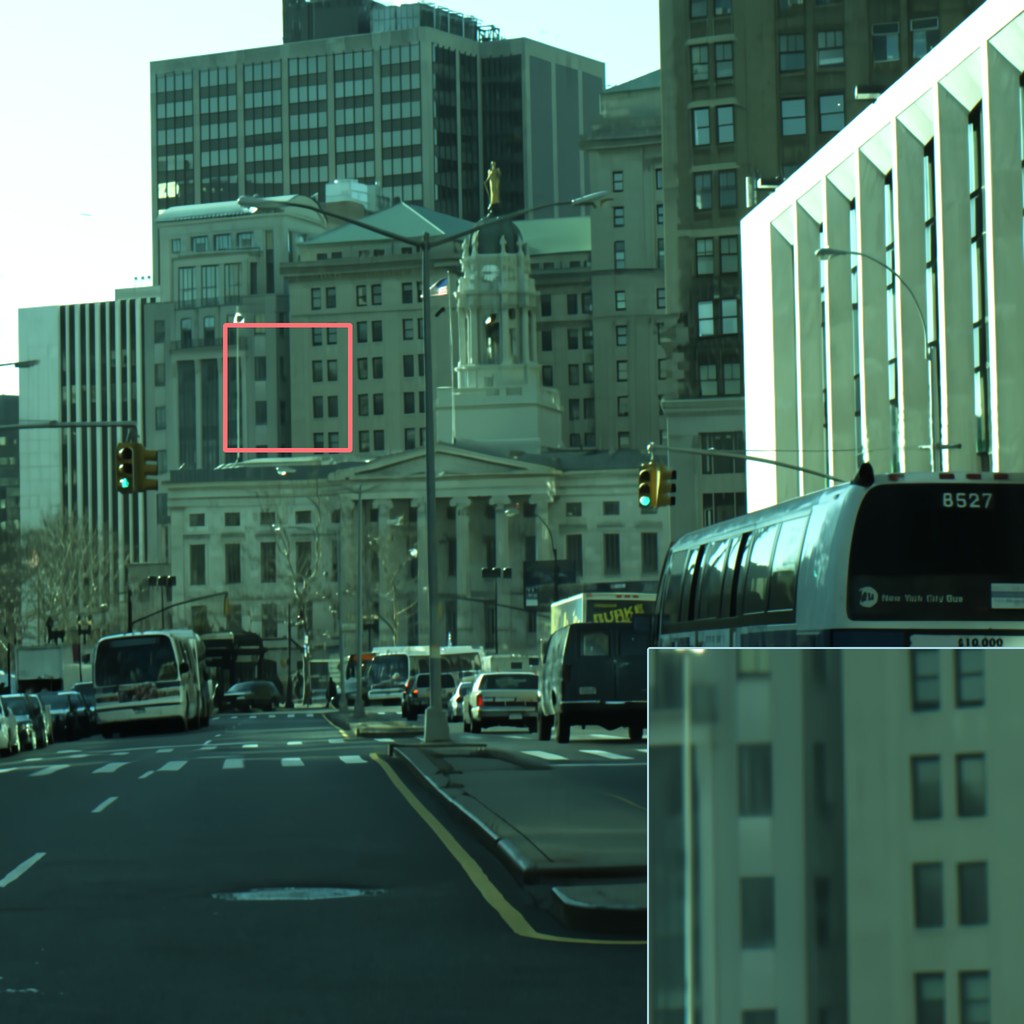} & 
         \includegraphics[width=0.245\linewidth]{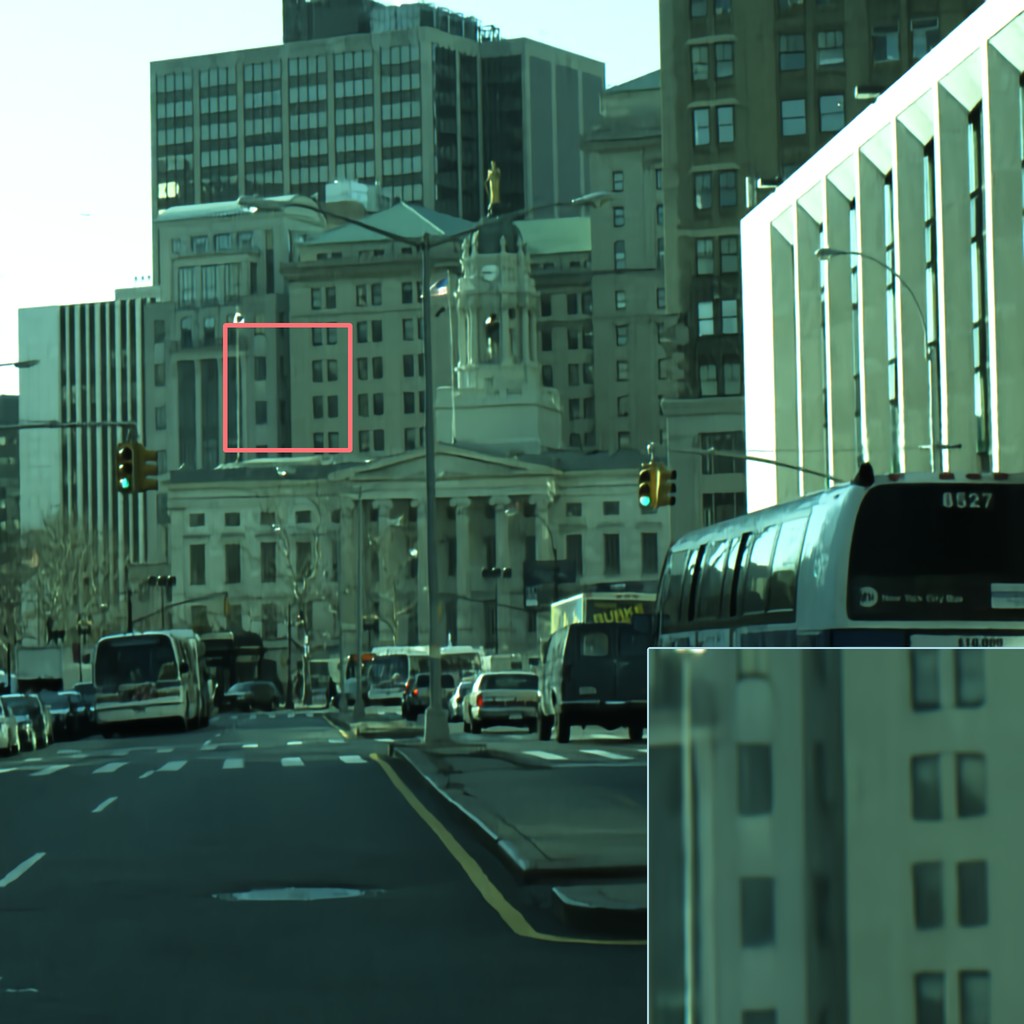}
         \tabularnewline
         LR Input & Team Samsung & EffectiveSR & RBSFormer~\cite{jiang2024rbsformer} \\
         
         \includegraphics[width=0.245\linewidth]{figs/test-samples/gt_177.png} & 
         \includegraphics[width=0.245\linewidth]{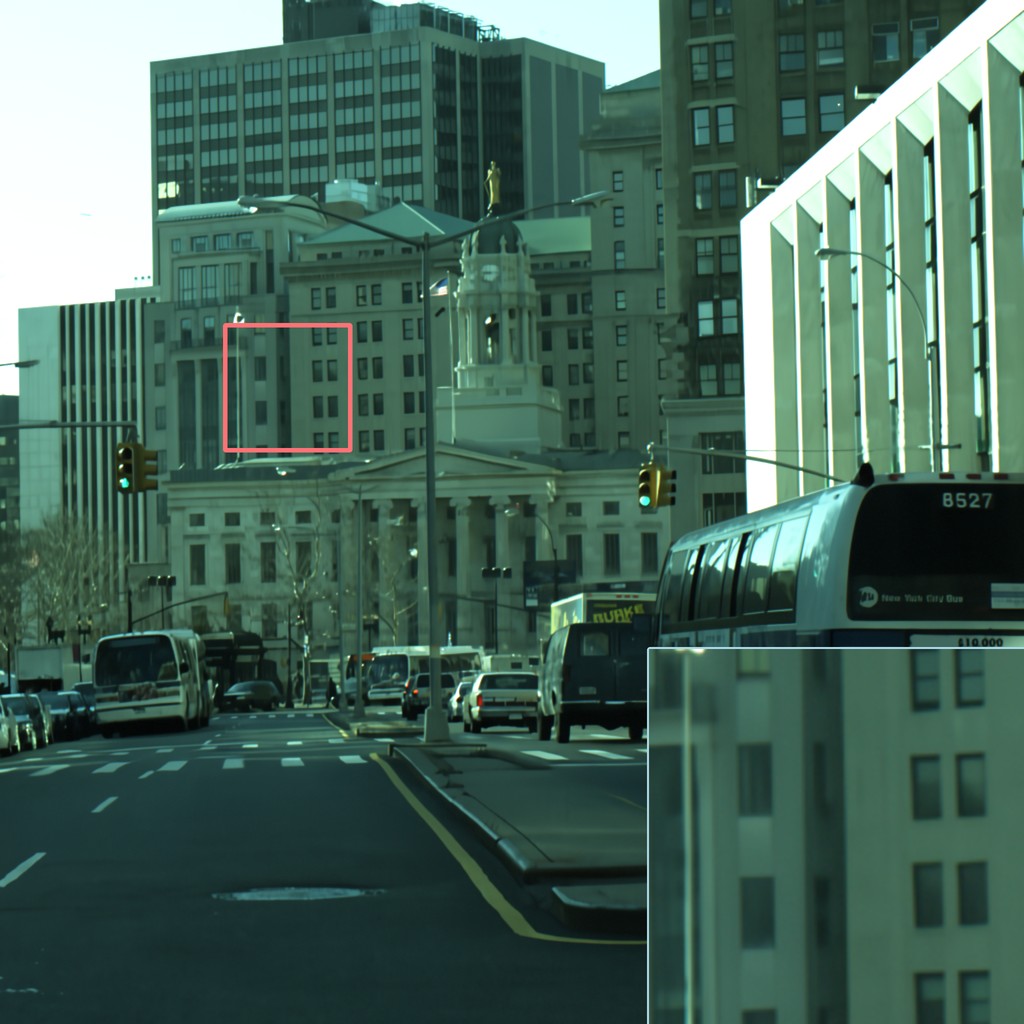} & 
         \includegraphics[width=0.245\linewidth]{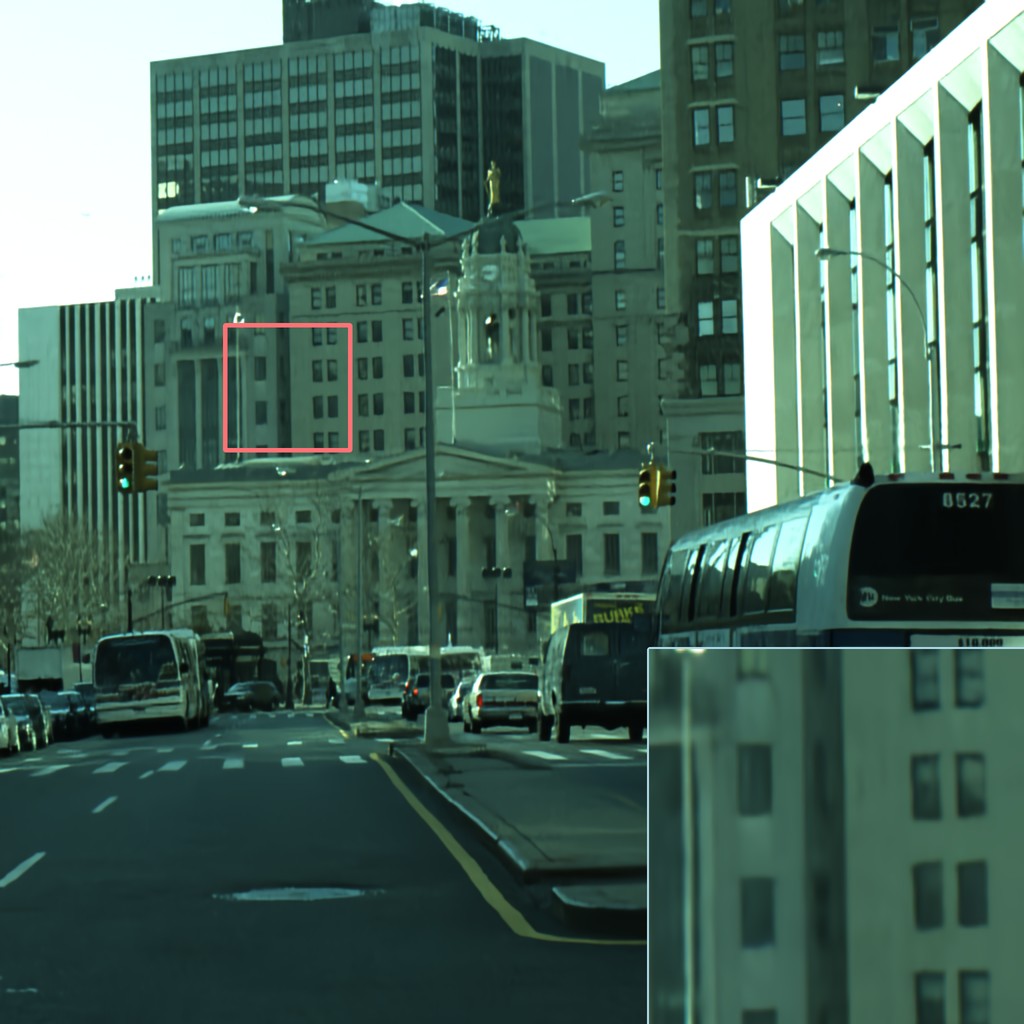} & 
         \includegraphics[width=0.245\linewidth]{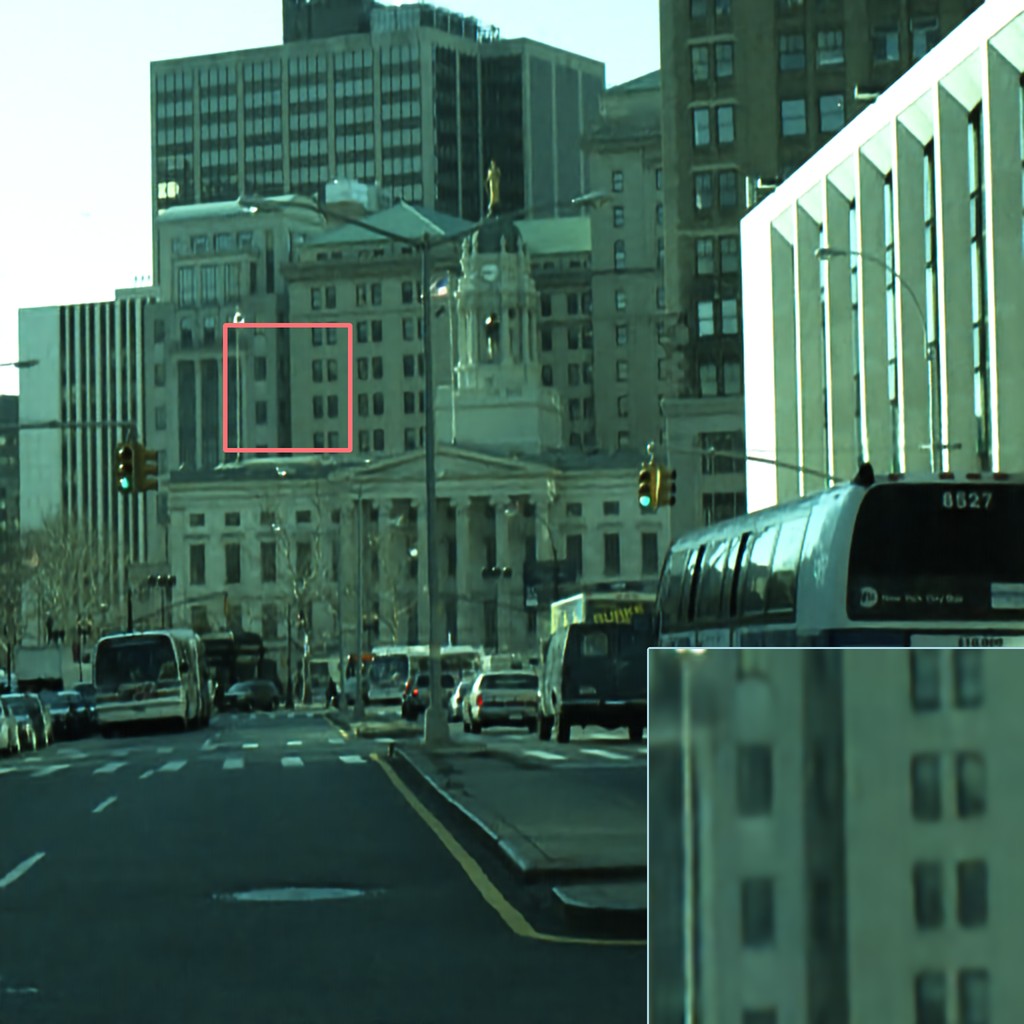}
         \tabularnewline
         HR Ground-truth & BSRAW~\cite{conde2024bsraw} & SwinFSR & SAFMN FFT \\
         % CROPS
         \includegraphics[width=0.245\linewidth]{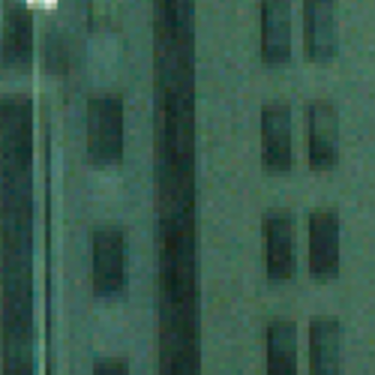} & 
         \includegraphics[width=0.245\linewidth]{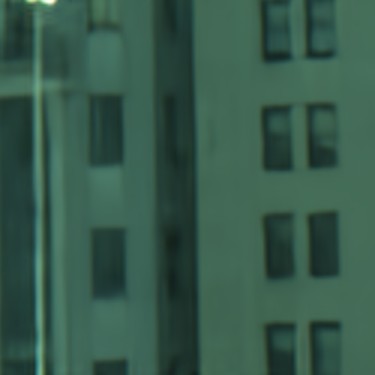} & 
         \includegraphics[width=0.245\linewidth]{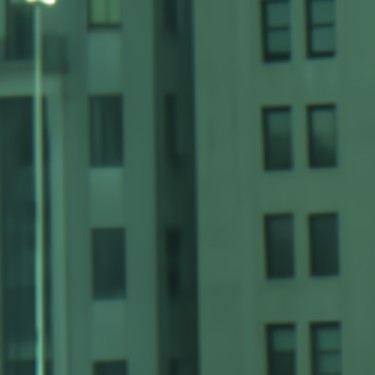} & 
         \includegraphics[width=0.245\linewidth]{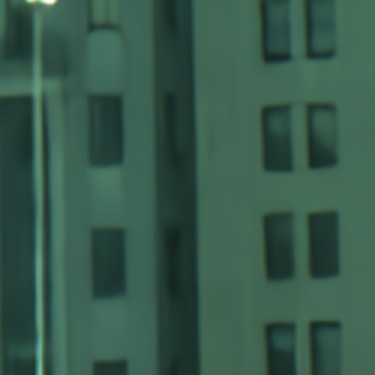}
         \tabularnewline
         LR Input & Team Samsung & EffectiveSR & RBSFormer~\cite{jiang2024rbsformer} \\
         
         \includegraphics[width=0.245\linewidth]{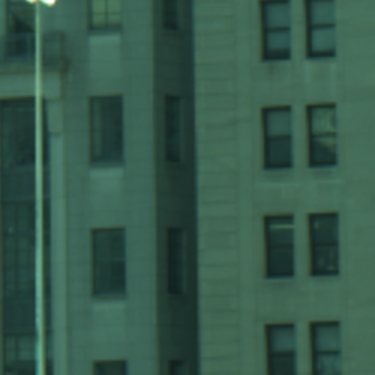} & 
         \includegraphics[width=0.245\linewidth]{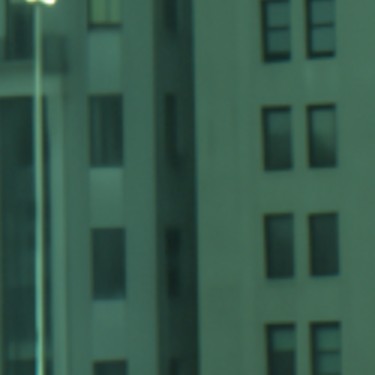} & 
         \includegraphics[width=0.245\linewidth]{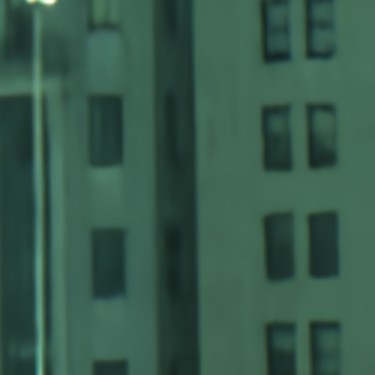} & 
         \includegraphics[width=0.245\linewidth]{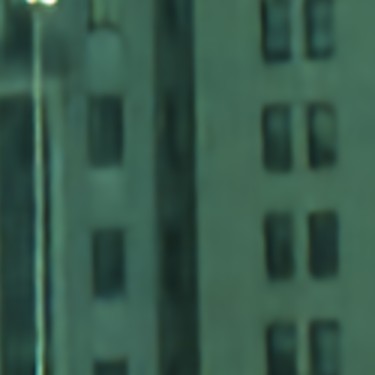}
         \tabularnewline
         HR Ground-truth & BSRAW~\cite{conde2024bsraw} & SwinFSR & SAFMN FFT
    \end{tabular}
    \caption{Visual comparison using the \textbf{NTIRE 2024 RAW Image Super-Resolution Challenge} testing set (\texttt{177.npz}). The HR resolution RAW images have $1024\times1024$ resolution and 4-channels (RGGB Bayer pattern). RAW images are visualized using bilinear demosaicing, gamma correction and tone mapping.}
    \label{fig:results3}
    \end{figure*}

In \cref{tab:benchmark} we provide the challenge benchmark. Besides fidelity metrics such as PSNR and SSIM, we also provide relevant implementation details of each method. The methods can greatly improve the RAW images quality and resolution, even in the case of full-resolution 12MP images as output. 
We provide detailed visual comparisons in \cref{fig:results1}, \cref{fig:results2} and \cref{fig:results3}. All the proposed methods are able to increase the resolution and details of the RAW images while reducing blurriness and noise. Moreover, there are not detectable color artifacts.

We can conclude that (synthetic) RAW image super-resolution can be solved similarity to RAW denoising. However, more realistic downsampling remains an open challenge.

\vspace{-3mm}

\paragraph{Acknowledgements}
This work was partially supported by the Humboldt Foundation. We thank the NTIRE 2024 sponsors: Meta Reality Labs, OPPO, KuaiShou, Huawei and University of W\"urzburg (Computer Vision Lab).

\newpage
\section{Challenge Methods and Teams}
\label{sec:teams}

%In the following sections we describe the best challenge solutions. Note that the method descriptions were provided by each team as their contribution to this survey. 

%Samsung
\subsection{Dual Stage RAW SR with Focal Pixel Loss}

%%%%%%%%%%% PLEASE FILL THIS INFORMATION WITH YOUR TEAM'S DETAILS
%%%% IF THE INFORMATION DOES NO MATCH THE FACTSHEET, WE WILL NOT INCLUDE IT

\begin{center}

\vspace{2mm}
\noindent\emph{\textbf{Team Samsung MX,SRC-B}}
\vspace{2mm}

\noindent\emph{Jianxing Zhang~$^1$,
Jia Li~$^1$,
Fan Wang~$^1$,
Xiaopeng Li~$^1$,
Zikun Liu~$^1$,
Hyunhee Park~$^2$,
Sejun Song~$^2$,
Changho Kim~$^2$}

\vspace{2mm}

\noindent\emph{
$^1$ Samsung Research China - Beijing (SRC-B)\\
$^2$ Samsung MX(Mobile eXperience) Business\\
}

\end{center}
%%%%%%%%%%%%%%%%%%%%%%%%%%%%%%%%%%%%%%%%%%%%%%%%%%%%%%%%%%%%%%%%%%
%%%%%%%%%%%%%%%%%%%%%%%%%%%%%%%%%%%%%%%%%%%%%%%%%%%%%%%%%%%%%%%%%%

%The goal of the RAW Image Super Resolution Challenge ~\cite{conde2024bsraw} is to up-sample a 4-channel RAW image with the degradation of blur and/or noise. In the challenge, we mainly adopted a two-stage neural network, which inspired by the work Restormer~\cite{zamir2022restormer} and NAFNet~\cite{chen2022simple}. We use the dataset provided by the organizer for training. It only contains 1000+ RAW images from DSLR cameras. We found more degradation methods for the RAW images to generate more low-resolution degraded samples for training. 

Team Samsung MX,SRC-B is introducing a two-stage network for RAW Image Super Resolution. The solution is using the divide-and-conquer strategy, with the first stage tasked with recovering the image structure from the low resolution degraded RAW image, and the second stage aiming at recovering the maximum amount of details, offering a refined reconstruction. 
Moreover, the team is further extending existing methods for synthetic data generation, studying further into hardware specific RAW image degradations, proposing new definitions for the relevant device-specific noise profiles and new blur kernels aligning with typical real-world scenarios. They propose a randomized degradation model, simulating different interactions between the observed simulated defects. 

Finally, Team Samsung MX,SRC-B is proposing a novel Focal Pixel Loss, which was proven through the performance improvements during the model fine-tuning stage.

\begin{figure}[t]
    \centering
    \includegraphics[width=\linewidth]{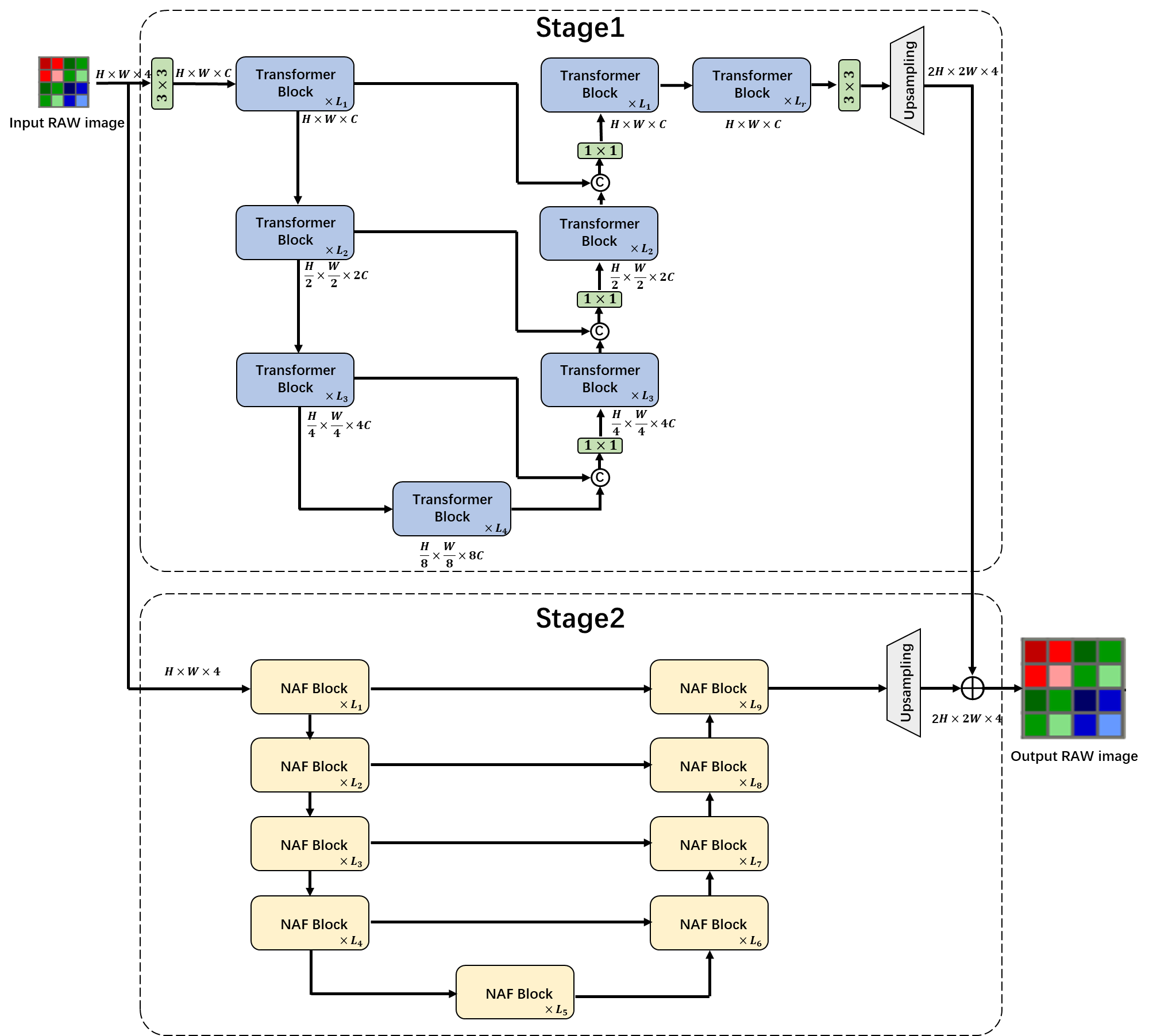}
    \caption{Dual-Stage RAWSR Framework proposed by Team Samsung MX,SRC. Best viewed in electronic version.}
    \label{fig:team2}
\end{figure}

As shown in \cref{fig:team2}, the network structure mainly includes two stages. The first stage mainly draws on Restormer~\cite{zamir2022restormer}, whose main role is to restore the main content of raw images. The second stage mainly uses a NAFNet~\cite{chen2022simple} based design, whose main role is to restore more details on the basis of the first stage of recovery. 

The training procedure accounts for the dual stage design of the proposed method. The first stage of the training procedure follows the optimization of the parameters corresponding to the first stage of the model. In the second stage of the training procedure, the optimized parameters are frozen, with the parameters of the second stage starting being refined, for optimum specialization. The final estimation is then performed using both sets of optimized parameters (Restormer and NAFNet models).

one of the main details of the solution is the proposal of a novel training objective, based on the characteristics of the input data. The proposition of the Focal Loss (see \cref{eq:focalloss}) is a solution for the observed imbalance in terms of highly affected pixels ratio, given the non-unifor effect of the signal degradation function . Therefore, the Focal Pixel Loss (FPL) is introducing exponential penalties to those pixels characterized by a large signal shift.

\begin{equation}
    FPL(\hat{I}, I)=-D(\hat{I}, I)^\gamma \log_{10} D(\hat{I}, I)
    \label{eq:focalloss}
\end{equation}

In \cref{eq:focalloss}, $D(., .)$ is a standard L-norm distance between the restored image $\hat{I}$ and the reference image $I$,  and $\gamma$ is an adjustable factor, controlling the strength of the penalty. 

\paragraph{Implementation details}

The model is trained solely on the data provided by the challenge organizers. It only contains more than 1000 RAW images from various DSLR camera sensors. The dataset was augmented using standard image augmentation techniques and simulations of the image degradation pipeline. The training procedure is a dual stage operation, optimizing the model stages sequentially. 
The optimization technique used is  the AdamW~\cite{decoupled} optimizer ($\beta_1=0.5$, $\beta_2=0.999$, weight decay $0.0001$) with the cosine annealing strategy, where the learning rate gradually decreases from the initial learning rate $5\times10^{-5}$ to $1\times10^{-7}$ for $5\times10^5$ iterations. 
The model goes through a pre-train optimization phase, base don the $L_1$ loss. During the fune-tune phase, the objective is set to the  Focal Pixel Loss, with the initial learning rate being set to $5\times10^{-6}$. The training batch size is set to $4$ and patch size is $384$. Horizontal/vertical flipping and rotation are used for data augmentation. All experiments are conducted on A100 GPUs.

%Xiaomi Inc., China
\subsection{EffectiveSR for RAW Images}

%%%%%%%%%%% PLEASE FILL THIS INFORMATION WITH YOUR TEAM'S DETAILS
%%%% IF THE INFORMATION DOES NO MATCH THE FACTSHEET, WE WILL NOT INCLUDE IT

\begin{center}

\vspace{2mm}
\noindent\emph{\textbf{Team XiaomiMMAI}}
\vspace{2mm}

\noindent\emph{Zhijuan Huang$^1$,
Hongyuan Yu$^1$,
Cheng Wan$^2$,
Wending Xiang,
Jiamin Lin$^1$,
Hang Zhong$^1$, 
Qiaosong Zhang$^1$, 
Yue Sun$^1$, 
Xuanwu Yin$^1$, 
Kunlong Zuo$^1$}

\vspace{2mm}

\noindent\emph{
$^1$Xiaomi Inc.
$^2$Georgia Institute of Technology
}

\end{center}
%%%%%%%%%%%%%%%%%%%%%%%%%%%%%%%%%%%%%%%%%%%%%%%%%%%%%%%%%%%%%%%%%%
%%%%%%%%%%%%%%%%%%%%%%%%%%%%%%%%%%%%%%%%%%%%%%%%%%%%%%%%%%%%%%%%%%

The solution proposed by Team XiaomiMMAI is a dual branch network based on HAT~\cite{chen2023activating}, adopting re-parameterization~\cite{ding2021repvgg} during training, using the additional parameters to fully exploit the potential of the method. They are introducing, a task-by-task and step-by-step training method for RAW Image Super-Resolution to simultaneously address three tasks: denoising, deblurring, and 2$\times$ Super Resolution. 

%RAW Image SR (RAWISR) involves a 2x scale expansion on raw images, which contain unknown noise and blur. Therefore, RAWISR needs to simultaneously address three tasks: denoising, deblurring, and 2x SR, and the model needs a large number of parameters to learn these tasks. 

To address the limitation given by the low number of samples offered for training, Team XiaomiMMAI converts the RAW images into an RGB images and performs combined data enhancement on the set of produced RGB image, using random rotations, flips, color changes, brightness changes, random blur, etc.. Then, the set of enhanced RGB images are translated back to the RGGB RAW domain, with the processed data used to train~\cite{brooks2019unprocessing} the proposed solution. 

\begin{figure}[t]
    \centering
    \includegraphics[width=0.3\textwidth]{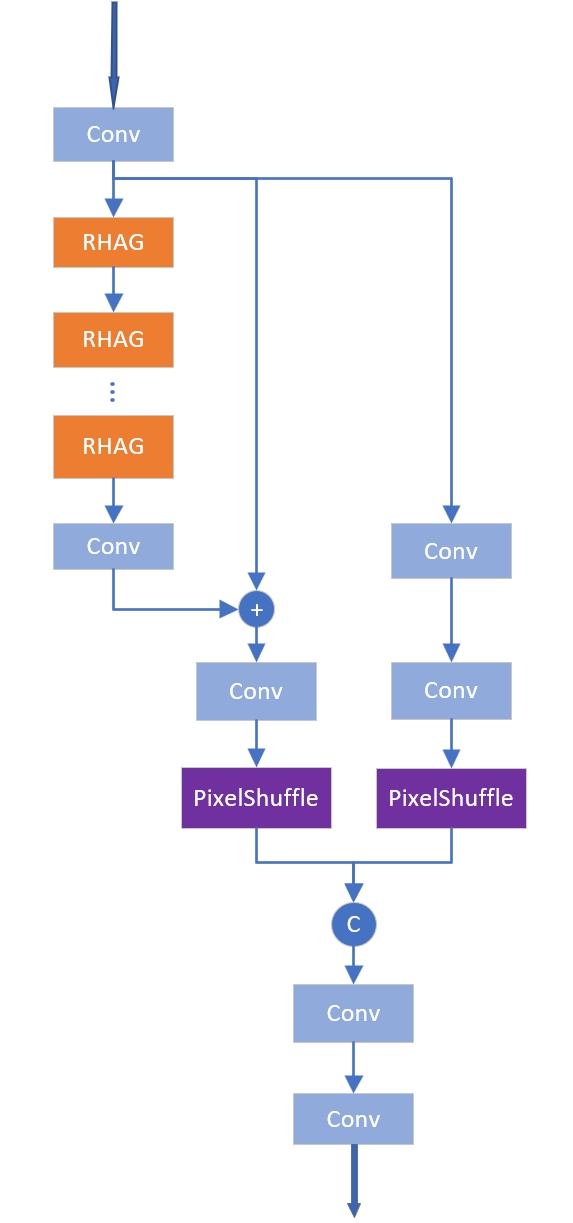}
    \caption{DB-HAT model proposed by Team XiaomiMMA}
    \label{fig:Dual_branch_network}
\end{figure}

The model proposed by Team XiaomiMMAI is inspired by HAT \cite{chen2023hat}, with the architecture being optimized for the RAW Image Super Resolution task. The optimized dual-branch network structure (DB-HAT) is shown in \cref{fig:Dual_branch_network}. The introduced Step-by-step and task-by-task training method for RAWISR further enhances the performance level achieved by their solution.

\textbf{Step-by-step}: To accelerate training and achieve good performance,  Team XiaomiMMAI adopted a strategy where each sub-task, including the final joint optimization, is trained based on a pyramid image representation. Initially, the model is trained on small scale images (64$\times$64), gradually increasing the resolution of the image patches to 128$\times$128 and 256$\times$256.

\textbf{Task-by-task}: Team XiaomiMMAI divided RAWISR into three sub-tasks: denoising, deblurring, and 2x SR. Initially, they start by training for RAW image denoising, followed by the connected tasks of deblurring and 2$\times$ Super Resolution. Finally, a joint optimization procedure is applied for entire network to produce the final estimator.

\begin{figure}[t]
    \centering
    \includegraphics[width=0.45\textwidth]{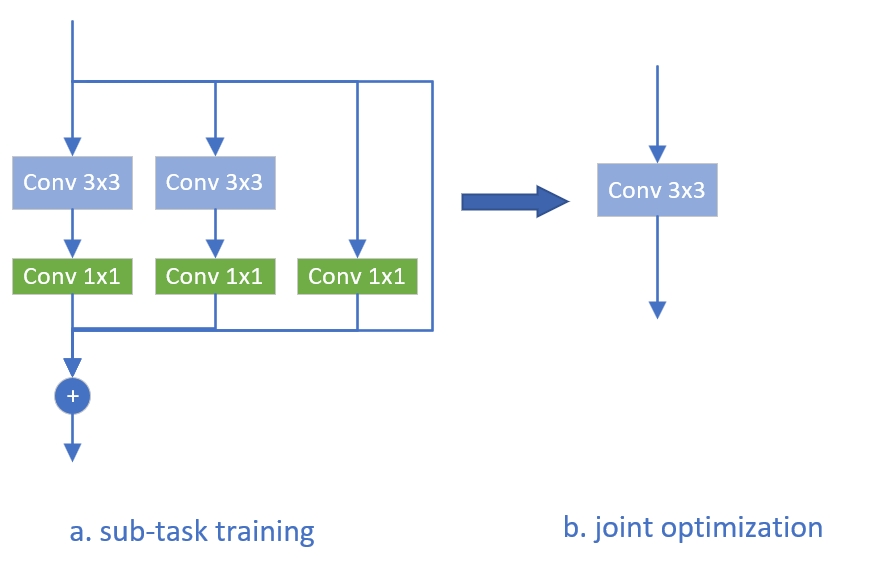}
    \caption{RepConv used by Team XiaomiMMA}
    \label{fig:RepConv}
\end{figure}

In the process of training denosing and deblurring, Team XiaomiMMAI used the RepConv re-parameterization technique on the final stage of the proposed  DB-HAT, to improve the visual image quality of the task. The reparameterizable convolution block (RepConv) is shown in Fig~\ref{fig:RepConv}. 

\paragraph{Implementation details}

The dataset used for three sub-task training consists of 1000+ RAWs, and the data augmentation methods can refer to the previous section for details. In the final stage of joint optimization of the entire network, Team XiaomiMMAI used the provided 1000+ dataset instead of the augmented dataset. The learning rate is initialized at 4 × $10^{-4}$ and decays according to the CosineAnnealing strategy during the training of three sub-tasks. The network undergoes training for a total of 2x$10^5$ iterations, with the L2 loss function being minimized as the trainign objective of the Adam optimizer.

Subsequently, finetuning is executed for two iterations, using the L2 loss and SSIM loss functions, with an initial learning rate of 5 × $10^{-5}$ for 2x$10^5$ iterations. All experiments are conducted with the PyTorch 2.0 framework on 8 A100 GPUs. 

%USTC604
\begin{figure*}[!ht]
    \centering
    \includegraphics[width=\linewidth]{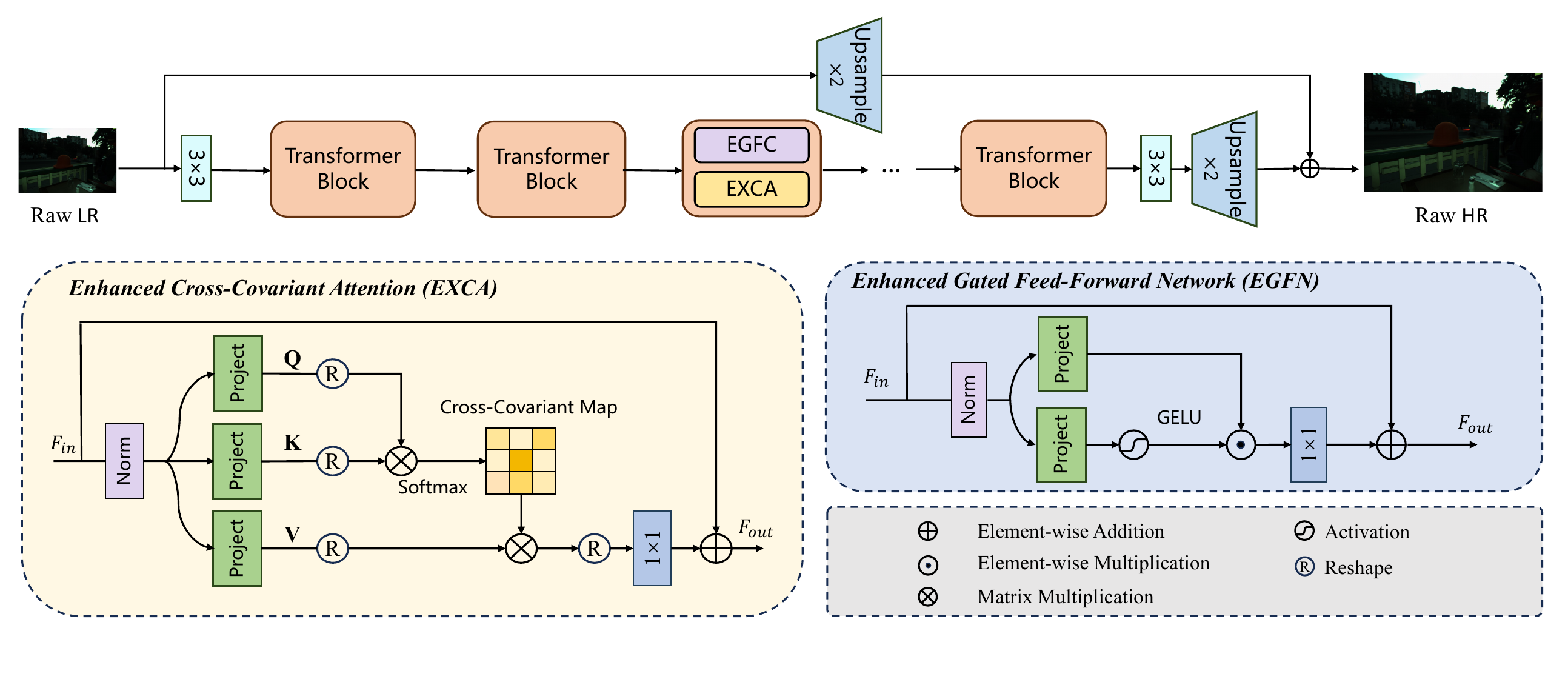}
    \caption{The RBSFormer~\cite{jiang2024rbsformer} Framework proposed by Team USTC604.}
    \label{fig:team4}
\end{figure*}

\subsection{RBSFormer: Enhanced Transformer Network for Raw Image Super-Resolution}

%%%%%%%%%%% PLEASE FILL THIS INFORMATION WITH YOUR TEAM'S DETAILS
%%%% IF THE INFORMATION DOES NO MATCH THE FACTSHEET, WE WILL NOT INCLUDE IT

\begin{center}

\vspace{2mm}
\noindent\emph{\textbf{Team USTC604}}
\vspace{2mm}

\noindent\emph{Senyan Xu,
Siyuan Jiang,
Zhijing Sun,
Jiaying Zhu}

\vspace{2mm}

\noindent\emph{University of Science and Technology of China}

\vspace{2mm}

\noindent{\emph{Contact: \url{a804235820@gmail.com}}}

\end{center}
%%%%%%%%%%%%%%%%%%%%%%%%%%%%%%%%%%%%%%%%%%%%%%%%%%%%%%%%%%%%%%%%%%
%%%%%%%%%%%%%%%%%%%%%%%%%%%%%%%%%%%%%%%%%%%%%%%%%%%%%%%%%%%%%%%%%%

Team USTC604 proposed a transformer framework for raw image super-resolution, with a design based on the transformer block proposed in Restormer\cite{zamir2022restormer} (see \cref{fig:team4}), solution that excels in capturing long-range pixel interactions by applying self-attention across channels. 

The solution used the data provided in the NTIRE 2024 RAW Image Super Resolution challenges, with the degradation pipeline described in \cite{conde2024bsraw}. 

For a 4-channel RGGB RAW image patch of size 224$\times$224, the computational cost of the model proposed by Team USTC604 amounts to 14.6 GFLOPS, being characterized by a number of 3.31 trainable parameters. On a consumer-grade gaming GPU, the NVIDIA RTX4090Ti, the forward pass needed of a full resolution image estimation needs 650 ms, following the limitations of the used backbone, as one of the computationally expensive solutions proposed in the image restoration field.  

The software characteristic to the performed experiments is based on the PyTorch 1.8 framework, with the experiments being performed on NVIDIA RTX4090Ti devices.  The trainign procedure is based on the Adam optimizer with the decay parameters parameters $\beta_1 = 0.9$ and $\beta_2 = 0.99$. The initial learning rate is $3 \times 10^{-4}$ and changes with Cosine Annealing scheme to $1 \times 10^{-7}$, with the training procedure covering 120K iterations in a time-frame of around 10 hours. We refer the reader to the author's paper RBSFormer~\cite{jiang2024rbsformer} for more details.

%McMaster

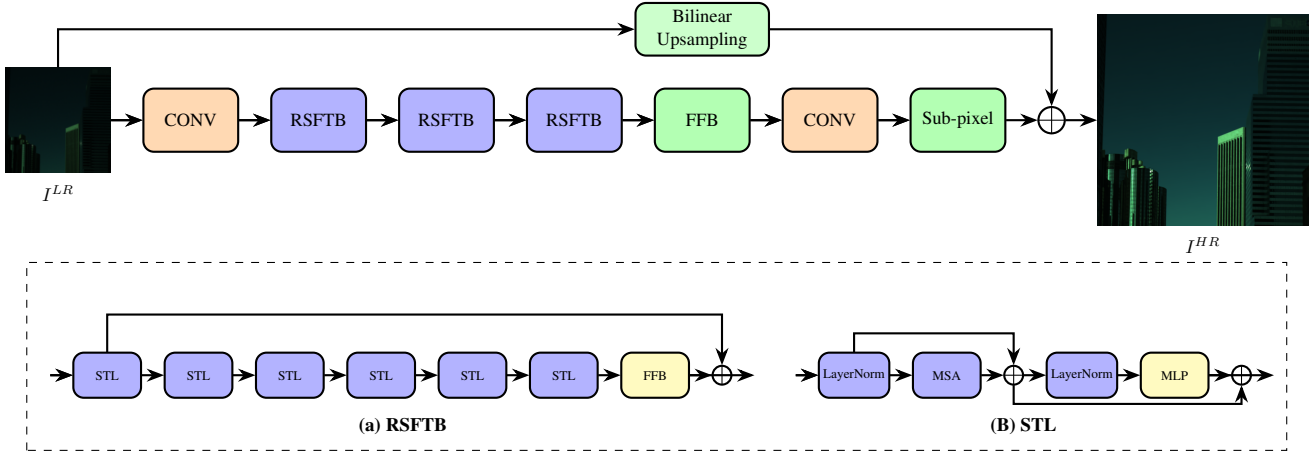
\begin{figure*}[!ht]
    \centering
\begin{tikzpicture}[
    start chain=going right,
    node distance=6mm and 6mm,
    layer/.style={draw, thick, rounded corners, minimum height=1.2cm, minimum width=1.8cm, align=center, on chain, join=by arrow},
    arrow/.style={-Stealth, thick},
    bilinear/.style={draw, thick, rounded corners, fill=green!20, align=center, minimum height=1cm, minimum width=2.5cm, text width=2.3cm},
    scale=0.7,
    transform shape,
    circuit symbol/.style={draw, thick, circle, minimum size=5mm, on chain, append after command={
        [very thick, every circuit symbol/.try] (\tikzlastnode.north) edge (\tikzlastnode.south)
        (\tikzlastnode.east) edge (\tikzlastnode.west)
        }
    },
    image/.style={inner sep=0pt, on chain}
]

% Define the style for different types of layers
\tikzset{
    input_raw/.style={layer, fill=red!30},
    conv/.style={layer, fill=orange!30},
    rcam/.style={layer, fill=yellow!30},
    subpixel/.style={layer, fill=green!30},
    others/.style={layer, fill=blue!30},
}

% Nodes
\node [image] (raw) {\includegraphics[width=2cm,height=2cm]{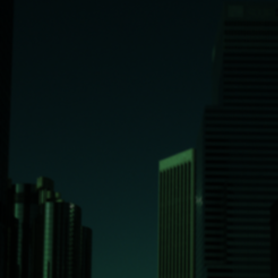}}; % Adjust size as needed
\node [conv, right=of raw] (conv1) {CONV};
\node [below=of raw, yshift=0.5cm]{$I^{LR}$};
\node [others, right=of conv1] (resft1) {RSFTB};
\node [others, right=of resft1] (resft2) {RSFTB};
\node [others, right=of resft2] (resft3) {RSFTB};
\node [subpixel, right=of resft3] (FFB) {FFB};
\node [conv, right=of FFB] (conv2) {CONV};
\node [subpixel, right=of conv2] (subpixel) {Sub-pixel};
\node [circuit symbol] (circuit) [right=of subpixel] {};
\node [image, right=of circuit] (out) {\includegraphics[width=4cm,height=4cm]{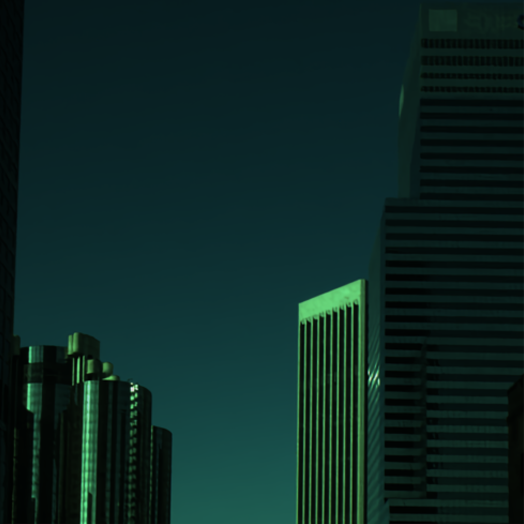}}; % Adjust size as needed
\node [below=of out, yshift=0.5cm]{$I^{HR}$};
% Bilinear Upsampling Block
\node [bilinear, above=of $(resft3.north)!0.5!(conv2.north)$] (bilinear) {Bilinear Upsampling};

% Connections
\draw [arrow] (raw) -- (conv1);
\draw [arrow] (conv1) -- (resft1);
\draw [arrow] (resft1) -- (resft2);
\draw [arrow] (resft2) -- (resft3);
\draw [arrow] (resft3) -- (FFB);
\draw [arrow] (FFB) -- (conv2);
\draw [arrow] (conv2) -- (subpixel);
\draw [arrow] (subpixel) -- (circuit);
\draw [arrow] (circuit) -- (out);

% Skip Connection
\draw [arrow] (raw.north) -- ++(0,0.5) |- (bilinear.west);
\draw [arrow] (bilinear.east) -- ++(1.5, 0) -| (circuit);

\end{tikzpicture}
\begin{tikzpicture}[
    start chain=going right,
    node distance=6mm and 6mm,
    layer/.style={draw, thick, rounded corners, minimum height=1.2cm, minimum width=1.8cm, align=center, on chain, join=by arrow},
    arrow/.style={-Stealth, thick},
    bilinear/.style={draw, thick, rounded corners, fill=green!20, align=center, minimum height=1cm, minimum width=2.5cm, text width=2.3cm},
    scale=0.5,
    transform shape,
    circuit symbol/.style={draw, thick, circle, minimum size=5mm, on chain, append after command={
        [very thick, every circuit symbol/.try] (\tikzlastnode.north) edge (\tikzlastnode.south)
        (\tikzlastnode.east) edge (\tikzlastnode.west)
        }
    },
    image/.style={inner sep=0pt, on chain}
]

% Define the style for different types of layers
\tikzset{
    input_raw/.style={layer, fill=red!30},
    conv/.style={layer, fill=orange!30},
    rcam/.style={layer, fill=yellow!30},
    subpixel/.style={layer, fill=green!30},
    others/.style={layer, fill=blue!30},
}

% Nodes
\node[on chain] (start) {};
\node [others ] (stl1) {STL};
\node [others ] (stl2) {STL};
\node [others ] (stl3) {STL};
\node [others ] (stl4) {STL};
\node [others ] (stl5) {STL};
\node [others ] (stl5) {STL};
\node [rcam ] (stl6) {FFB};
\node [circuit symbol] (circuit) [right=of stl6] {};
\node[on chain] (end) {};
\draw [arrow] (stl1.north) -- ++(0, 1)-| (circuit);
\draw [arrow] (stl6) -- (circuit);
\draw [arrow] (start) -- (stl1);
\draw [arrow] (stl6) -- (end);
\node[align=center, below] at (9.5,-1) {\Large\textbf{(a) RSFTB}};

\node[on chain] (start) {};
\node [others ] (stl1) {LayerNorm};
\node [others ] (stl2) {MSA};
\node [circuit symbol ] (stl4) {};
\node [others ] (stl5) {LayerNorm};
\node [rcam ] (stl6) {MLP};
\node [circuit symbol] (circuit) [right=of stl6] {};
\node[on chain] (end) {};
\draw [arrow] (stl1.north) -- ++(0, 0.5)-| (stl4);
\draw [arrow] (stl4.south) -- ++(0, -0.5)-| (circuit);
\draw [arrow] (stl6) -- (circuit);
\draw [arrow] (start) -- (stl1);
\draw [arrow] (stl6) -- (end);
\draw [arrow] (stl2) -- (stl4);
\node[align=center, below] at (26,-1) {\Large\textbf{(B) STL}};
\draw[dashed] (-0.5,-2) rectangle (33,3);
\end{tikzpicture}
    \centering
    \caption{Main branch of the framework proposed by Team McMaster. (a) Residual Swin Fourier Transformer Block (RSFTB)
    \\(b) Swin Transformer Layer (STL).}
    \label{fig:my_diagram}
\end{figure*}

\subsection{SwinFSR Raw Image Super Resolution}

%%%%%%%%%%% PLEASE FILL THIS INFORMATION WITH YOUR TEAM'S DETAILS
%%%% IF THE INFORMATION DOES NO MATCH THE FACTSHEET, WE WILL NOT INCLUDE IT

\begin{center}

\vspace{2mm}
\noindent\emph{\textbf{Team McMaster}}
\vspace{2mm}

\noindent\emph{Liangyan Li,
Ke Chen,
Yunzhe Li,
Yimo Ning, 
Guanhua Zhao,
Jun Chen
}

\vspace{2mm}

\noindent\emph{McMaster University}

\vspace{2mm}

\noindent{\emph{Contact: \url{lil61@mcmaster.ca}}}

\end{center}
%%%%%%%%%%%%%%%%%%%%%%%%%%%%%%%%%%%%%%%%%%%%%%%%%%%%%%%%%%%%%%%%%%
%%%%%%%%%%%%%%%%%%%%%%%%%%%%%%%%%%%%%%%%%%%%%%%%%%%%%%%%%%%%%%%%%%

%Image Super-Resolution received plenty attention in the computer vision research community. Given the wide availability of the sRGB images, most of the conventional image super-resolution methods\cite{wang2021uformer,chen2022simple,zamir2020cycleisp,zamir2022restormer}, are designed with a significant dependency on the sRGB image representation. However, it is important to note that sRGB images are result of a highly nonlinear transform, whose input is the original raw data. The transform is characterized by information loss, as the sensor specific representation is translated to a standard representation. 

%Moreover, even with the standardized sRGB image representation, the transformation from the sensor-specific RAW representation implies camera-specific Image Signal Processing (ISP) pipelines. This inconsistency, combining both hardware and software factors, makes the design process of an universal Super Resolution algorithm, working on both sRGB and RAW domains, a complex problem. Such models, when generalized, may underperform for specific camera systems. 

%Tailoring unique models to individual camera systems for optimal performance incurs significant costs. Even if a model would benefit from learning on data characterized by the same hardware-specific properties that would be met during deployment, given the wide range of camera products, this strategy becomes non-feasible.

Team McMaster proposed an algorithm that considers multiple acquisition sensors, accounting for various image signal degradations induced by hardware limitations. The model is trained and learned directly from the 4-channel RAW data with an enhanced degradation pipeline. With a broader variance of noise during the degradation process, the solution demonstrates increased robustness, efficiently producing high-quality images from degraded inputs, thereby enhancing overall performance on the official datasets. 
Their approach works directly with 4-channel RGGB RAW images after a designed degradation process. The architecture is a hybrid model which integrates SwinFSR \cite{chen2023swinfsr} with simple CNN layers. The model was trained and validated only on the official datasets \cite{conde2024ntire_raw}. 

\begin{comment}
The key point for end-to-end raw image super resolution is the degradation process of original high resolution images.
\begin{equation}\label{equation1}
    y=(x  \otimes k){\downarrow s} + n
\end{equation}

As a general degradation model, \cref{equation1} assumes that the low resolution(LR) image $y$ is corresponding to a High Resolution(HR) image $x$, on which the blur kernel $k$ is applied, then downsampled by the scale factor $s$.  Finally, the result is affected by the hardware-specific noise sources, modeled by the profile of the noise factor $n$.     
\end{comment}

The team utilized the noise model, blur kernel,  and degradation model as demonstrated in BSRAW~\cite{conde2024bsraw}.

Their approach of adding noise is inspired by the strategy deployed in DiT \cite{peebles2022scalable} that gradually adds Gaussian noise as a form of degradation in the forward process. Diffusion models\cite{hodenoising2020,peebles2022scalable} have been proposed for various imaging tasks, including image super-resolution, demonstrating their capability for end-to-end training to transform pure Gaussian noise into meaningful data representations. They investigate the applicability of gradual magnitude Gaussian noise, as utilized in diffusion models, for addressing the degradation process inherent in Raw Image Super-Resolution tasks. Team McMaster adopted the additive noise model from \cite{peebles2022scalable}, exposing the input RAW images to noise using the forward diffusion definition described in \cite{peebles2022scalable}. In the forward diffusion process, the input data undergoes a degradation process the iterative addition of Gaussian noise across 1000 discrete steps. In the backward denoising process of DiT, the model is optimized for the forward diffusion process, maximizing the data likelihood via the variational lower bound.% , and defined as:
%     \[x_0: q(x_{t} | x_{0}) = \mathcal{N}(x_t;\sqrt{\bar{\alpha}_t}x_0,(1-\bar{\alpha}_t)\mathbf{I})\]
%     This approach sequentially adds noise to images, starting from noise towards reconstructing the original image. The result after added noise can be formulated as: $x_t = \sqrt{\bar{\alpha}_t} x_0 + \sqrt{1 - \bar{\alpha}_t} \epsilon_t $ \cite{peebles2022scalable} where $\bar{\alpha}_t$ is from the pre-trained model and $\epsilon_t \sim \mathcal{N}(0,\mathbf{I})$

The architectural configuration of the Team McMaster proposed model is depicted in \cref{fig:my_diagram}.
% Transformers \cite{vaswani2017attention} and CNNs are popular deep learning architectures used for raw image super-resolution tasks.
In the proposed method, a SwinFSR-based design \cite{keswinfsr2023} performs RAW domain image feature extraction, combined with feature upsampling, matching the size of high-resolution images via a complex convolution operator. SwinFSR builds on the success of SwinIR \cite{liang2021swinir}, with an additional data modality given by the Frequency Domain Knowledge, through the FFT image representation \cite{chen2023swinfsr}. This proves to be a superior strategy, combining spatial and spectral features as a way to balance the local information of the spatial domain, and the global information accessed through the spectral representation. It introduces a novel cross-attention module for efficient information exchange between the two modalities and adapts to rectangular input patches for flexibility. 

%With shared weights and strategic information exchange, alongside optimized training strategies, SwinFSR achieves a superior performance-complexity tradeoff, outperforming the baseline on stereo-super resolution tasks. 

For the proposed model, only the feature extraction branch of SwinFSR is deployed.

\paragraph{Implementation details}
The solution was optimized solely on the NTIRE 2024 official challenge data \cite{conde2024ntire_raw}, using the proposed Development Phase submission set for validation. This dataset contains 1064 4-channel DSLR-specific RGGB RAW images for training, and an additional 40 raw images set for validation. The images were pre-processed, applying white-black level correction, then being normalized to the unit interval. Since the degraded low-resolution raw images are characterized by low quality, with a considerable level of details being lost, a data augmentation technique is applied, to improve the training procedure in terms of stability, convergence, and the achieved performance level. The strategy combines simple horizontal or vertical flips with channel shifts and mixup augmentations. The training objective is based on the  $L_1$ loss.

%NUDT\_RSR
\subsection{Spatially-Adaptive Feature Modulation for RAW Super-Resolution}

%%%%%%%%%%% PLEASE FILL THIS INFORMATION WITH YOUR TEAM'S DETAILS
%%%% IF THE INFORMATION DOES NO MATCH THE FACTSHEET, WE WILL NOT INCLUDE IT

\begin{center}

\vspace{2mm}
\noindent\emph{\textbf{Team NUDT RSR}}
\vspace{2mm}

\noindent\emph{Jinyang Yu,
Kele Xu,
Qisheng Xu,
Yong Dou}

\vspace{2mm}

\noindent\emph{National University of Defense Technology, Computer Dept., Changsha, China}

\vspace{2mm}

\noindent{\emph{Contact: \url{a804235820@gmail.com}}}

\end{center}
%%%%%%%%%%%%%%%%%%%%%%%%%%%%%%%%%%%%%%%%%%%%%%%%%%%%%%%%%%%%%%%%%%
%%%%%%%%%%%%%%%%%%%%%%%%%%%%%%%%%%%%%%%%%%%%%%%%%%%%%%%%%%%%%%%%%%

\begin{figure*}[!ht]
    \centering
    \includegraphics[trim={0 1cm 0 0},clip, width=0.9\textwidth]{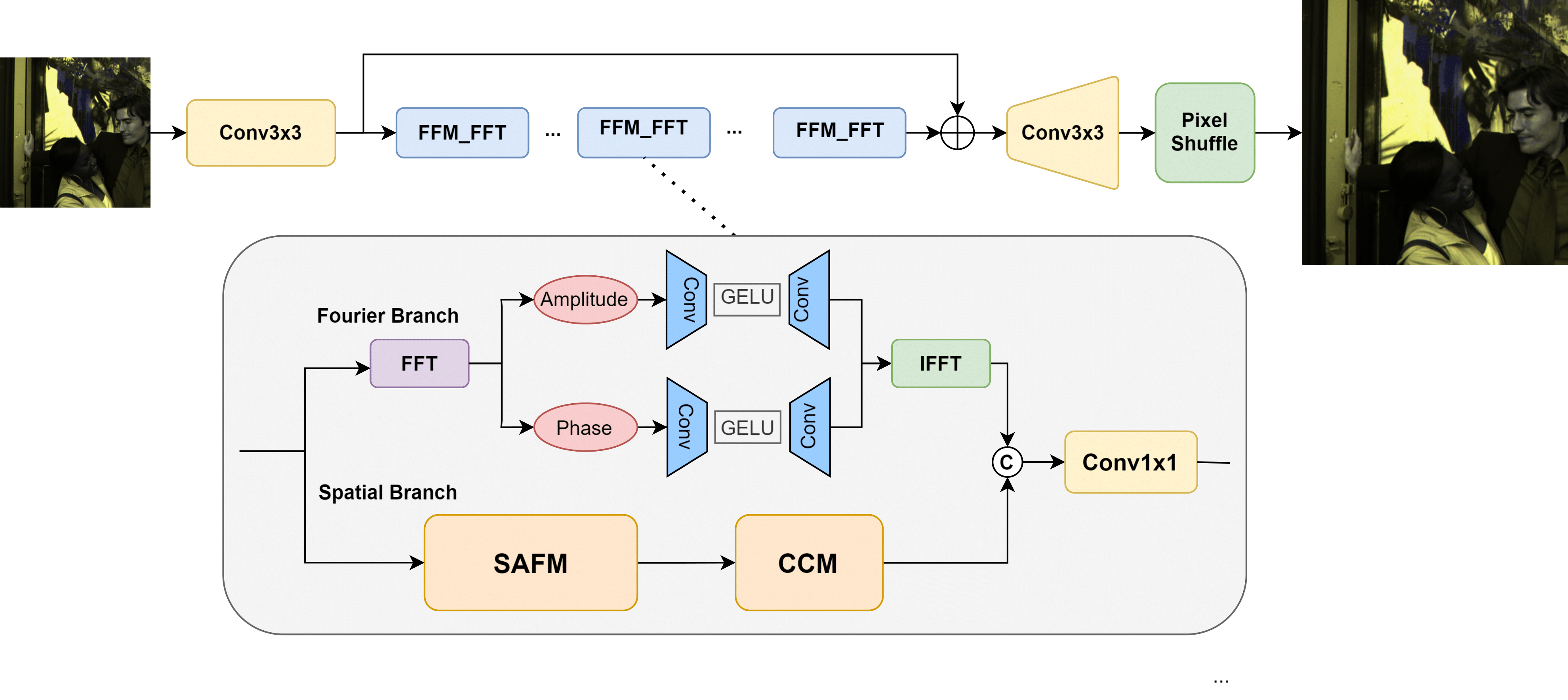}
    \caption{The overall network architecture proposed by Team NUDT RSR. The proposed FFM\_FFT block is a two-branched dolution, with the Fourier branch extracting the amplitude and phase, guiding the global-local feature mixing performed in the spatial branch, through SAFM and CCM blocks\cite{10376717}.}
    \label{fig:team1}
\end{figure*}

The solution proposed by Team NUDT RSR addresses three major components of the proposed challenge, extending on the degradation pipeline, model design, and model supervision, achieving a significant performance level in terms of restoration fidelity. 

 For the image signal degradation pipeline, the considered degradations include diverse blur kernels, exposure defects, image downsampling, finally coupled with a noise model characteristic to real-world raw image data. 
 
 Following \cite{li2023embedding}, the proposed solution is benefitting from the amplitude and phase components, added to the SAFMN~\cite{10376717} backbone. the model is fusing frequency domain and spatial domain information, for global-local level feature mixing.
 
 Moreover, the knowledge distillation is deployed to the described solution, using a NAFNet~\cite{chen2022simple} teacher, with multiple complexity level feature supervision. Finally, we apply a progressive training strategy, gradually increasing the patch size at each stage to accommodate larger test inputs.

%The Raw Image Super-resolution challenge aims at upsampling a degradation-unknown 4-channel Raw Image by a factor of two with efficient methods. In our method, the work is divided into 3 parts, including the degradation pipeline, the network designing and the training strategy.

\textbf{Degradation pipeline:}
    Inspired by \cite{9607421} and \cite{conde2024bsraw}, in order to enable the model to learn real degradation information, the Bayer pattern RAW images are cropped, following by degrading the RAW signal in a sequence of operations composed of multiple blurring operations, exposure compensations, downsampling, and hardware specific added noise.

    The first step is represented by a blurring operator, based on randomly generated Gaussian blur, generalized Gaussian blur, with a plateau-shaped distribution, and their an-isotropic version. The augmented PSF kernels provided by starter kit are also considered. All kernel sizes are ranging from $7\times7$ to $25\times25$.

    Then, the pipeline continues with linear adjustment for image exposure.  As \cite{conde2024bsraw} discussed, to simulate the artifacts caused by underexposure and overexposure, the pipeline implements exposure adjustment by linearly scaling the image. The adjustment factor is tuned to the [-0.25, 0.25] interval, applied in the unit interval normalized images.

    Next, the image suffers further downsampling, considering different downsampling kernels, including bicubic interpolation, bilinear interpolation, and an average-pooling operator. To build multi-scale training pairs, the input image is either upsampled or downsampled to a random size first, and rescaled back to half of the original size as for 2$\times$ super-resolution tasks.

    On the downscaled image, heteroscedastic Gaussian noise \cite{conde2024bsraw} is then applied, followed by the practical shot-read noise \cite{conde2024bsraw} for different exposure levels. An image with higher exposure factor in step 2 is more likely to get noised by heteroscedastic noise, and the shot-read noise for low-light images.

    The last step of the degradation pipeline is a second blurring operator. To expand the degradation space like the high-order degradation model \cite{9607421}, a random operation based on a set of second blurring kernels (same kernels considered in the first step), characterized by smaller standard deviations, is applied in the final stage. 

\begin{table}
    \centering
    \resizebox{\linewidth}{!}{
    \begin{tabular}{l c c c}
         \toprule
         Method & Params.~[M] & FLOPS~[G] & PSNR~[dB]  \\
         \midrule
         NAFNet~(Large) & 116 & 255 & 41.76  \\
         NAFNet~(Small) & 0.290 & 14.08 & 40.78 \\
         \midrule
         SAFMN & 0.229 & 59.06  & 41.20 \\
         SAFMN\_FFT & 0.272 & 67.29  & 41.81 \\
         \bottomrule
    \end{tabular}
    }
    \caption{A comparison between the NAFNet, vanilla SAFMN and the  method proposed by Team NUDT RS. The large version of NAFNet uses [2, 2, 4, 8] encoding blocks, [2, 2, 2, 2] decoding blocks for each stage, and 12 middle blocks, the width is set to 64. While the smaller version use width 32, with both [2, 1] configuration for encoding and decoding blocks and 1 middle block. The vanilla SAFMN has dim=36, ffn\_scale=2 configuration, and 8 main blocks, which are the same to teh poposed SAFMN\_FFT. All flops are calculated with input size $1\times4\times512\times512$.}
    \label{tab:nudt_ablation}
\end{table}

\vspace{-2mm}

\paragraph{Network}
  According to \cite{li2023embedding}, the Fourier spectrogram of an image has similar amplitude to its downsampled one's, while the phase is related to the noise observed in the acquired image signal. Although it is designed for low light image enhancement tasks, Team NUDT RS started with the observation that blind RAW Image super-resolution can also benefit from refining these two components of an low resolution image.
  
  Thus, the Team NUDT RS proposed model is represented in  \cref{fig:team1}, where the input image is encoded by a $3 \times 3$ convolution layer for shallow feature extraction, and the FFM\_FFT blocks are used for deep feature extraction. Following \cite{li2023embedding}, the main blocks are divided into spatial branches and Fourier branches. In each block, the input is simultaneously sent to both branches, then the processed features of the branches are fused. After a residual connection, a final pixel-reshuffle operator is used to upscale the feqature set to the resolution of the reference image \cite{7780576}.

  In the spatial branch, the cross domain communication is performed through efficient Feature Mixing Module (FMM) blocks of SAFMN~\cite{10376717}. A FMM block consists of a SAFM block and a CCM block. SAFM splits the channels into different parts, fuses their features at different scale levels and obtain attention map after GELU activation. Then, the original input multiplies the attention map. The CCM block consists of a $3 \times 3$ convolution layer and a $1 \times 1$ convolution layer, which works as a channel mixer to capture local context information.
  
  In the Fourier branch, an image is transformed to frequency map by Fast Fourier Transform (FFT) operator to get amplitude and phase component, which are later processed by two $1 \times 1$ convolution layers with GELU activations respectively.  Next, these refined components are combined to a new frequency map, and the inverse FFT its used to transfer back to the spatial domain. Features from different branches are concatenated and fused by another $1 \times 1$ convolution. The proposed model uses  8 such blocks with width 36. To avoid expensive computations on high dimensions, the SFT layers in the final reconstruction stage are removed \cite{li2023embedding}. This results in an efficient estimator, with the total parameter count being 272.068 K for the proposed solution. The team presents an ablation study in \cref{tab:nudt_ablation}.

\vspace{-2mm}
\paragraph{Training strategy}
A two-stage optimizing strategy is applied. All the training is based on the development dataset and there are no external datasets used.
  
Firstly, a large version of NAFNet~\cite{chen2022simple} model is trained as a teacher network, with increased complexity degradations on patch size 128.  Then this NAFNET model is used to apply knowledge distillation, as part of the optimization technique used for the proposed solution. To reduce the performed computations, the knowledge  does not rely on distances computed between multi-level feature sets, but rather on statistics defining the feature see \cite{10.1007/978-3-030-20890-5_34}.
  
Secondly, to accommodate the high resolution test images, the student model is progressively finetuned on increased resolution patches. The training patch size increases from 128, 256, 352 to 448.

The optimization objective is based on the L! distance simultaneously applied in the spatial and Fourier domains~\cite{NEURIPS2022_6e60a902}. The total loss is defined in \cref{eq:team1loss}.

\begin{equation}
     \mathcal{L}_p = \left\|I_1 - I_2\right\|_1
\end{equation}
\begin{equation}
     \mathcal{L}_f = \left\|\mathcal{F}(I_1) - \mathcal{F}(I_2)\right\|_1 
\end{equation}
\begin{equation}
    \mathcal{L}_{kd} = \frac{1}{|D|}\sum_{i\in D}{\left\| \mathcal{G}(\mathcal{N}_i(I_1)) - \mathcal{G}(\mathcal{N}_i(I_2)) \right\|_{c} }
\end{equation}
\begin{equation}
    \mathcal{L}_{total} = \mathcal{L}_p + \lambda\mathcal{L}_f + \mu\mathcal{L}_{kd}
    \label{eq:team1loss}
\end{equation}

$I_1, I_2$ denotes the restored image prediction and the corresponding reference image. $\mathcal{F}(\cdot)$ is the FFT operator, $\mathcal{N}_i(\cdot)$ is the feature extracting operator determined by the network, where $i\in D$ is a set of the intermediate layers and $\mathcal{G}(\cdot)$ is defined to the square of the channel-wise mean of intermediate features. The $\left\|\cdot\right\|_{c}$ represents Charbonnier loss. $\lambda$ and $\mu$ are used to control the weights of different components. The finetuning stage only uses the $\mathcal{L}_p$ and $\mathcal{L}_f$ loss terms.

\vspace{-2mm}

\paragraph{Implementation details}
The experiments are based on the Pytorch framework for implementation. The NAFNet teacher model uses a width 64, with 2, 2, 4, 8 encoder blocks and 2, 2, 2, 2 decoder blocks at each stage. The middle block num is set to 12. The teacher module is trained for 600000 iterations. The learning rate is set to 2e-4, and halves for every 100000 steps after 300000 steps. 
The training procedure applied for the teacher network uses only the $\mathcal{L}_p$ loss. For knowledge distillation, the procedure starts with an initial learning rate warm-up stage, lasting for 30000 iterations. Then, a cosine annealing stage is applied, decreasing the learning rate to 1e-6. 

The FFT L1 loss is balanced with a weight $\lambda=0.05$, and the distillation loss with $\mu=0.1$. For the progressive finetuning, each training stage for the proposed model lasts for 200,000 iterations with learning rate 5e-5 that is reduced to 1e-6 with a cosine scheduler. Alongside the augmentation involved in the degradation pipeline, some traditional data augmentation methods, such as random cropping, flipping, and rotation are deployes to further diversify the training dataset. The testing phase deploys the geometric self-ensemble strategy \cite{8014885}, which averages 8 outputs corresponding to 8 augmented versions of the input image. All the experiments are conducted on a NVIDIA GeForce RTX 2080Ti GPU, using the AdamW optimizer with $\beta_1=0.99$, $\beta_2=0.9$. 

\newpage
%%%%%%%%%%%%%%%%%%%%%%%%%%%%%%%%%%%%%%%%%%%%%%%%%%%%%%
%%% VISUAL COMP

%%%%%%%%%%%%%%%%%%%%%%%%%%%%%%%%%%%%
%%%%%%%%%%%%%%%%%%%%%%%%%%%%%%%%%%%%
\newpage
%%%%%%%%%%%%%%%%%%%%%%%%%%%%%%%%%%%%
%%%%%%%%%%%%%%%%%%%%%%%%%%%%%%%%%%%%
\newpage
%%%%%%%%%%%%%%%%%%%%%%%%%%%%%%%%%%%%
%%%%%%%%%%%%%%%%%%%%%%%%%%%%%%%%%%%%

%%%%%%%%%%%%%%%%%%%%%%%%%%%%%%%%%%%%%%%%%%%%%%%%%%%%%%
%%%%%%%%% REFERENCES
\clearpage
{\small
\bibliographystyle{ieeenat_fullname}
\bibliography{refs}
}

\end{document}